%% file: main.tex
\documentclass[letterpaper, 10 pt, conference]{ieeeconf}  

\IEEEoverridecommandlockouts                              

\usepackage{graphicx} 
\usepackage{epsfig} 
\usepackage{mathptmx} 
\usepackage{times} 
\usepackage{amsmath} 
\usepackage{amssymb} 
\usepackage{xcolor}

\input{h2t_def}

\usepackage[capitalise]{cleveref}
\usepackage[per-mode=fraction]{siunitx}
\usepackage{subcaption}
\usepackage{import}
\usepackage{stfloats}
\usepackage[sort]{cite}
\usepackage{lipsum}
\usepackage{multicol}
\usepackage{multirow}
\usepackage{setspace}
\usepackage{soul}
\usepackage{balance}
\usepackage{svg}

\DeclareMathAlphabet{\mathcal}{OMS}{cmsy}{m}{n}

\graphicspath{{figures/}}

\usepackage{xfrac}
\usepackage{nicefrac}
\DeclareSIUnit{\Nm}{\newton\metre}
\DeclareSIUnit{\ms}{\milli\second}
\DeclareSIUnit{\dps}{\degree\per\second}
\DeclareSIUnit{\mmps}{\milli\metre\per\second}
\DeclareSIUnit{\mps}{\metre\per\second}
\DeclareSIUnit{\MPa}{\mega\pascal}
\DeclareSIUnit{\hPa}{\hecto\pascal}
\DeclareSIUnit{\kPa}{\kilo\pascal}

\usepackage[acronym]{glossaries}
\setacronymstyle{long-short}

\newacronym{EMG}{EMG}{electromyography}
\newacronym{sEMG}{sEMG}{surface electromyography}
\newacronym{PCB}{PCB}{printed circuit board}
\newacronym{IMU}{IMU}{inertial measurement unit}
\newacronym{FMG}{FMG}{force myography}
\newacronym{FSR}{FSR}{force sensing resistor}
\newacronym{SMP}{SMP}{surface muscle pressure}
\newacronym{IMP}{IMP}{intramuscular pressure}
\newacronym{PCA}{PCA}{principal component analysis}
\newacronym{PFDF}{PF/DF}{plantar-/dorsiflexion}
\newacronym{INEV}{IN/EV}{in-/eversion}
\newacronym{IRER}{IR/ER}{internal/external rotation}
\newacronym{RMSE}{RMSE}{root-mean-squared error}
\newacronym{MSE}{MSE}{mean-squared error}
\newacronym{R2}{R$^2$}{coefficient of determination}
\newacronym{ME}{ME}{mean error}
\newacronym{GM}{GM}{gastrocnemius medialis}
\newacronym{GL}{GL}{gastrocnemius lateralis}
\newacronym{TA}{TA}{tibialis anterior}
\newacronym{RF}{RF}{rectus femoris}
\newacronym{BF}{BF}{biceps femoris}
\newacronym{ST}{ST}{semitendinosus}
\newacronym{VM}{VM}{vastus medialis}
\newacronym{VL}{VL}{vastus lateralis}
\newacronym{CoM}{CoM}{center of mass}
\newacronym{RoM}{RoM}{range of motion}
\newacronym{GPR}{GPR}{Gaussian process regression}
\newacronym{SGPR}{SGPR}{sparse gaussian process regression}
\newacronym{RBF}{RBF}{radial basis function}
\newacronym{BPN}{BPN}{back propagation neural network }
\newacronym{LRCN}{LRCN}{long-term recurrent convolution network}
\newacronym{LSTM}{LSTM}{long short-term memory}
\newacronym{SVM}{SVM}{support vector machine}
\newacronym{SVR}{SVR}{support vector regression}
\newacronym{ANN}{ANN}{artificial neural network}
\newacronym{CNN}{CNN}{convolutional neural network}
\newacronym{TCN}{TCN}{temporal convolutional network}
\newacronym{NMS}{NMS}{neuromusculoskeletal}
\newacronym{MAD}{MAD}{median absolute deviations}
\newacronym{GRF}{GRF}{ground reaction force}
\newacronym{LOPO}{LOPO}{leave-one-participant-out}
\newacronym{LOVO}{LOVO}{leave-one-velocity-out}

\makeatletter

\def\ps@titlepagestyle{%
    \def\@oddfoot{\mycopyrightnotice}%
    \def\@oddhead{\conferencenotice}%
    \def\@evenfoot{}%
}

\def\conferencenotice{\begin{minipage}{\textwidth}
  {\vspace{-10mm} \scriptsize \centering \begin{singlespace} Author's version:\\ This file corresponds to the manuscript presented at the IEEE International Conference on Robotics and Automation (ICRA), May 2025. \end{singlespace}}
  \end{minipage}
}

\def\mycopyrightnotice{\begin{minipage}{\textwidth}
  {\scriptsize \begin{singlespace} \copyright~2025 IEEE. Personal use of this material is permitted. Permission from IEEE must be obtained for all other uses, in any current or future media, including reprinting/republishing this material for advertising or promotional purposes, creating new collective works, for resale or redistribution to servers or lists, or reuse of any copyrighted component of this work in other works.\end{singlespace}}
  \end{minipage}}

\title{\LARGE \bf Force Myography Based Torque Estimation \\ in Human Knee and Ankle Joints}

\author{Charlotte Marquardt\authorrefmark{1}\authorrefmark{3}, Arne Schulz\authorrefmark{1}, Miha De\v{z}man\authorrefmark{1}, Gunther Kurz\authorrefmark{2}, Thorsten Stein\authorrefmark{2} and Tamim Asfour\authorrefmark{1}\authorrefmark{3}
\thanks{\ackJuBotRIG\newline 
    \authorrefmark{1}High Performance Humanoid Technologies Lab, Institute for Anthropomatics and Robotics, Karlsruhe Institute of Technology (KIT), Germany\newline
    \authorrefmark{2}BioMotion Center, Institute of Sports and Sports Sciences, Karlsruhe Institute of Technology (KIT), Germany\newline
    \authorrefmark{3} Corresponding authors: {\tt\scriptsize \{charlotte.marquardt,asfour\}@kit.edu}}%
}

\begin{document}

\maketitle


\begin{abstract}
The online adaptation of exoskeleton control based on muscle activity sensing offers a promising approach to personalizing exoskeleton behavior based on the user's biosignals.
While \acrfull{EMG}-based methods have demonstrated improvements in joint torque estimation, \acrshort{EMG} sensors require direct skin contact and extensive post-processing. In contrast, \acrfull{FMG} measures normal forces resulting from changes in muscle volume due to muscle activity.
We propose an \acrshort{FMG}-based method to estimate knee and ankle joint torques by integrating joint angles and velocities with muscle activity data. We learn a model for joint torque estimation using \acrfull{GPR}.  
The effectiveness of the proposed \acrshort{FMG}-based method is validated on isokinetic motions performed by ten participants.
The model is compared to a baseline model that uses only joint angle and velocity, as well as a model augmented by \acrshort{EMG} data. 
The results indicate that incorporating \acrshort{FMG} into exoskeleton control can improve the estimation of joint torque for the ankle and knee joints in novel task characteristics within a single participant.
Although the findings suggest that this approach may not improve the generalizability of estimates between multiple participants, they highlight the need for further research into its potential applications in exoskeleton control.
\end{abstract}


\input{sections/introduction}

\input{sections/methods}

\input{sections/results}

\input{sections/discussion}

\input{sections/conclusion}


\section*{ACKNOWLEDGMENT}
The authors would like to thank Tim Wolk (High Performance Humanoid Technologies Lab, Karlsruhe Institute of Technology (KIT)) for his support in facilitating training and testing on multiple participants.


\clearpage
\bibliographystyle{IEEEtran}
\bibliography{IEEEabrv,HumanoidsGroup,ICRA2025}

\end{document}

%% file: h2t_def.tex
\usepackage[per-mode=fraction]{siunitx}
\usepackage{comment}
\usepackage[normalem]{ulem}
\usepackage{xspace}
\usepackage{xcolor}
\usepackage{lastpage}
\usepackage{fancyhdr}
\usepackage{soul}

\DeclareSIUnit\px{px}
\DeclareSIUnit\fps{fps}

\definecolor{OliveGreen}{RGB}{0,200,25}
\newcommand{\red}[1]{\textcolor{red}{#1}}

\newcommand{\darkgreen}[1]{\textcolor{OliveGreen}{#1}}

\newcommand{\etal}{et\,al.\xspace}

\newcommand{\ackJuBotRIG}{This work has been supported by the Carl Zeiss Foundation through the JuBot project and the German Federal Ministry of Education and Research (BMBF) under the Robotics Institute Germany (RIG).}

\newif\iffinal

\newcommand{\replaced}[2]{%
	\iffinal%
	#2%
	\else%
	\red{\ifmmode\text{\sout{\ensuremath{#1}}}\else\sout{#1}\fi}\darkgreen{#2}%
	\fi%
}
\newcommand{\removed}[1]{%
	\iffinal%
	\else%
	\red{\ifmmode\text{\sout{\ensuremath{#1}}}\else\sout{#1}\fi}%
	\fi%
}



%% file: sections/introduction.tex
\section{INTRODUCTION}

Lower limb exoskeletons are wearable devices designed to assist or augment mobility. While their design and control of these devices have traditionally focused on joint biomechanics, there is growing interest in incorporating muscle-level biomechanics to improve the effective interaction with the wearer’s musculoskeletal system.
Incorporating muscle-level biomechanics into exoskeleton design and control has the potential to overcome some of the current limitations, such as lack of personalization in control strategies, including the need to manually adjust control for each user or track fatigue or energy expenditure~\cite{mahdian_tapping_2023}.
Insights from muscle biomechanics research can enhance understanding of human movement and improve the effectiveness of exoskeleton control~\cite{wang_integral_2022}.

To incorporate muscle biomechanics into exoskeleton control, methods are needed to measure or estimate these biomechanics. 
Such methods should provide relevant biomechanical data in real-time during static and dynamic motion and must be compatible with the physical structure of the exoskeleton.
\Gls{EMG} is a widely recognized approach to capturing the electrical effects of muscle activity~\cite{joshi_classification_2013,taborri_gait_2016,jiang_emerging_2021}. 
However, ensuring high-quality \gls{EMG} signals requires extensive filtering and signal post-processing. 
Several factors can negatively affect signal quality, including electrode positioning on the muscle, electrode skin contact, and electrode displacement during muscle contraction.
In contrast, \gls{FMG} detects the mechanical phenomena associated with muscle contraction rather than electrical effects by measuring the normal forces resulting from the muscle volume change.
Consequently, it does not require direct skin contact, precise sensor placement on the muscle, and complex post-processing~\cite{castellini_wearable_2014,Marquardt2022}. 
Since only contact between the body and the exoskeleton is required to measure the interaction forces between both, force sensors can be integrated into exoskeleton cuffs, making FMG-based control of exoskeletons a promising approach.

\begin{figure}[b]
    \centering
        \includegraphics[width=0.75\columnwidth]{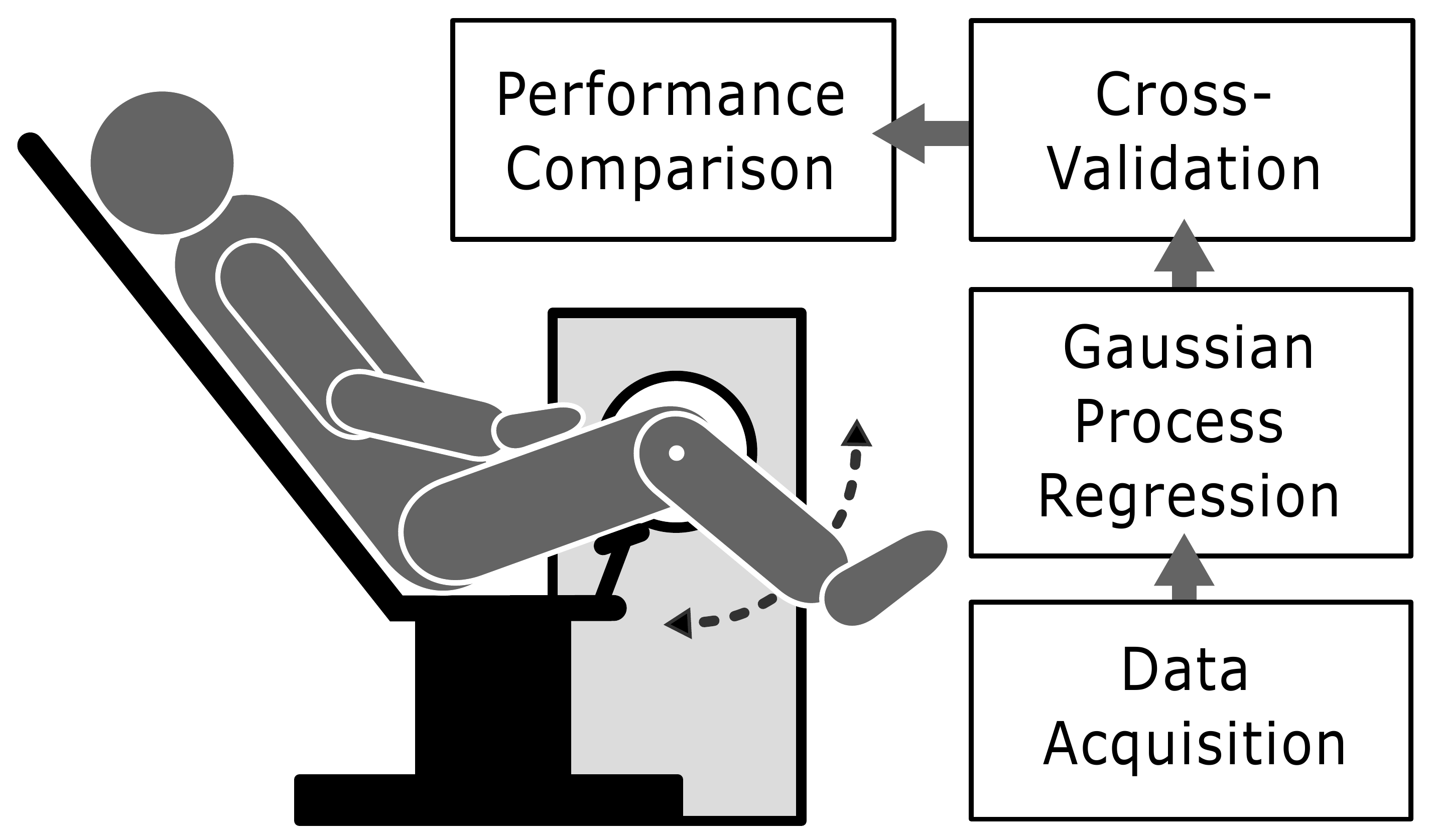}
 	\caption{Schematic overview of the process of data acquisition, regression, and evaluation.}
	\label{fig:overview_process}
\end{figure}

Estimating human joint torques allows unifying exoskeleton control and thus reduces the user effort~\cite{molinaro_estimating_2024}. 
In our previous work, we investigated the use of \gls{FMG} for exoskeletons using barometric pressure-based \gls{FMG} units~\cite{Marquardt2022, Marquardt2024}.
This paper extends our investigation of using \gls{FMG} to estimate joint torques of the knee and ankle joint based on the combination of the joint angle and velocity with muscle activity.
To do so, we learn a model for joint torque estimation using \gls{GPR}.
We consider the proposed \gls{FMG}-based approach for estimating joint torque an important initial step towards personalized exoskeleton control. 
We demonstrate the potential of the approach using data collected in a user study with ten participants performing isokinetic exercises, in which the velocity of the limb movement is maintained constant with varying resistance and muscle forces. 
To evaluate the \gls{FMG}-based approach, we compared our model to a baseline model that only uses joint angle and velocity, as well as a model augmented by \acrshort{EMG} data (\cref{fig:overview_process}).

The paper is organized as follows. \cref{sec:related_work} discusses current related work while \cref{sec:methods} describes the model used to estimate the joint torques, the user study conducted, the sensor setup used, and the processing of the recorded signals. The quality of the torque estimation and its validation results are presented in \cref{sec:results} and discussed in \cref{sec:discussion}. \Cref{sec:conclusion} concludes the paper.

\section{RELATED WORK}
\label{sec:related_work}

The introduction of joint torque estimation in exoskeleton control aims to provide momentous feedback in a closed-loop control system. This enables task-independent control, eliminating the need for discretization of human motion, resulting in task-dependent feedforward control trajectories~\cite{molinaro_estimating_2024}.
Using multiple sensing methods to capture muscle activity has created an opportunity to optimize human joint torque estimation during exoskeleton usage.
Previous work has explored a variety of \gls{EMG}-driven methods for actuation and estimation of joint torques in exoskeleton control, and intention prediction. 

\gls{EMG} methods combined with \gls{NMS} models enhance human-robot cooperation in exoskeletons by considering joint angles and muscle dynamics~\cite{ao_movement_2017,xu_development_2021}. 
This combination performs best during high muscle activation trials, but pairing it with \gls{ANN} showed superior performance with diverse training data~\cite{zhang_ankle_2021}. 
Recent hybrid \gls{NMS} models combined with \gls{ANN} and \gls{CNN} have been shown to outperform traditional \gls{NMS} models in torque estimation~\cite{zhang_ankle_2022,schulte_multi-day_2022}. 
Additionally, combining inertial sensors with \gls{EMG}-driven simulations supports the characterization of knee joint mechanics during walking~\cite{gurchiek_estimating_2019,gurchiek_wearables-only_2022}, while \gls{NMS} models have been shown to enable personalized torque estimations in the ankle joint using real-time \gls{EMG} data~\cite{durandau_neuromechanical_2022}.
Other \gls{EMG}-based methods use deep learning methods~\cite{hahn_neural_2008,siu_neural_2021,camargo_predicting_2022}, such as \glspl{TCN}~\cite{scherpereel_improving_2024}, \gls{CNN}~\cite{schulte_multi-day_2022} or \gls{ANN}~\cite{zhang_ankle_2022} directly to estimate lower limb joint torques and optimize its performance. 
Although these methods surpass traditional \gls{NMS} models, their effectiveness in a hybrid model with \gls{NMS} varies based on the specific model and application~\cite{schulte_multi-day_2022,zhang_ankle_2022}.
Proportional myoelectric control directly uses the muscle activity amplitude to control the torque output of the exoskeleton~\cite{hybart_preliminary_2022}. Studies showed that users of a proportional myoelectric controlled ankle exoskeleton maintained their normal joint biomechanics~\cite{hybart_neuromechanical_2023}, however, the effect on metabolic cost was limited during walking both on a treadmill and outdoors~\cite{hybart_metabolic_2023,hybart_gait_2023}. Moreover, these controllers measure only the resulting behavior of the muscle actions and thus do not fully capture the muscle and body mechanics~\cite{mahdian_tapping_2023}.
In general, these results show the potential to integrate \gls{EMG} signals, either directly or in advanced algorithms, to personalize joint torque estimation and torque control of exoskeletons.

On the other hand, \gls{FMG} signals have been extensively and successfully investigated in various wearable applications such as upper arm or hand motion classification and intention detection~\cite{islam_effective_2020,xiao_performance_2017,jiang_exploration_2017,belyea_fmg_2019}, lower limb gait phase or event detection~\cite{islam_novel_2022,jiang_ankle_2016} and ankle position classification~\cite{jiang_exploration_2018} often showing to outperform \gls{EMG}-based methods. 
Research indicates a nonlinear relationship between muscle surface deformation and torque, implying that muscle deformation could be an effective signal for non-invasive, real-time torque measurement~\cite{alvarez_surfacelevel_2024}.
Sakr et al.~\cite{sakr_estimation_2019} demonstrated promising results of using \gls{FMG} signals from the lower arm to estimate multi-directional isometric hand force/torque.
The feasibility of estimating lower-limb joint torques using \gls{FMG} signals has yet to be investigated and compared to methods based on EMG and kinematic sensors alone.

%% file: sections/methods.tex
\section{METHODS}
\label{sec:methods}

\begin{figure*}[htb]
    \vspace{1mm}
    \centering
    \includegraphics[width=0.99\textwidth]{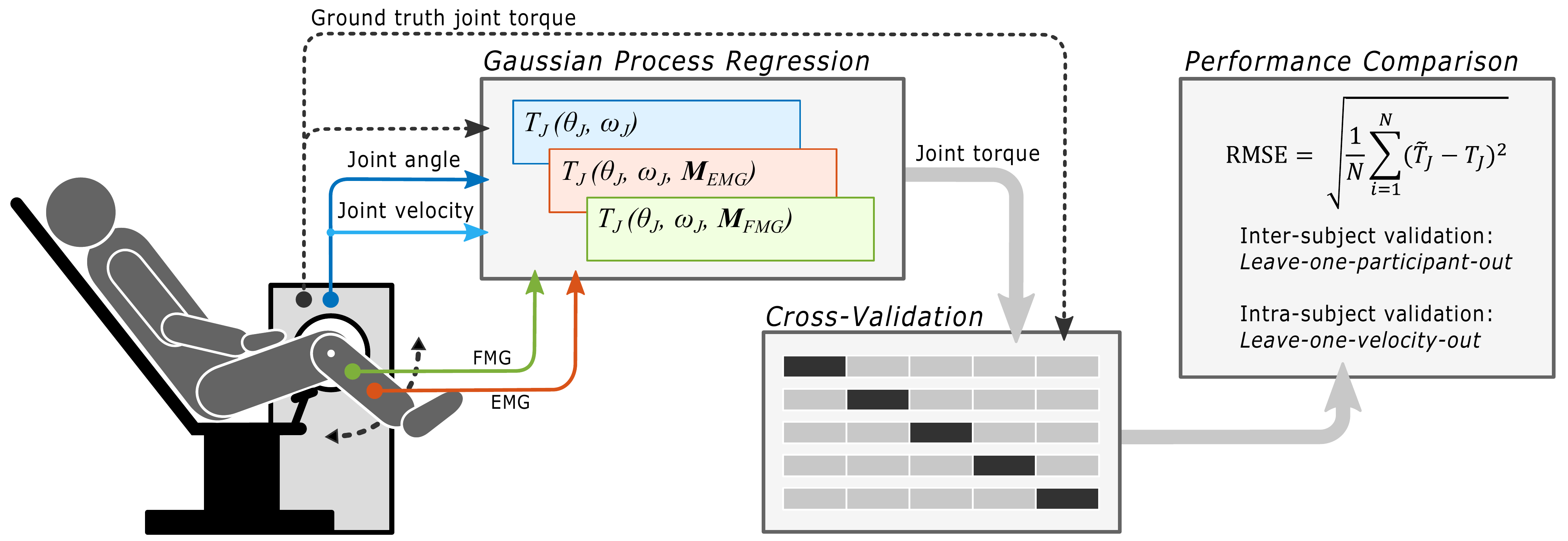}
    \caption{Schematic representation of the torque estimation and validation process for knee joint motion.}
    \label{fig:overview_data_pipeline}
\end{figure*}

Accurate biomechanical models of joint kinematics in combination with normal muscle forces are difficult to obtain. To learn the relationship between kinematics and muscle force, we propose a \gls{GPR} model.
This section describes the model and the user study, including the sensor setup used to provide data for the training and validation of the model.
A schematic overview of the entire torque estimation and validation process is given in~\cref{fig:overview_data_pipeline}.

\subsection{Joint Torque Estimation Model}

\gls{GPR} models are a kernel-based probabilistic and parametric supervised learning method for input-output mapping of empirical data that follows a joint Gaussian distribution~\cite{rasmussen_gaussian_2006}. In vector form, this can be expressed by
\begin{equation}
    f(\mathbf{x}) \sim \mathcal{GP}(m(\mathbf{x}),k(\mathbf{x},\mathbf{x}')),
    \label{eq:GPRmodel}
\end{equation}
where an observed outcome~$f(\mathbf{x})$ is estimated from an input~$\mathbf{x}$ by a Gaussian process with the mean function~$m$ and the covariance function~$k(\mathbf{x},\mathbf{x}')$. 
For the covariance function, the \gls{RBF}
\begin{equation}
    k(\mathbf{x}, \mathbf{x}') = \sigma  ^2 \exp(-\frac{| \mathbf{x} - \mathbf{x}' |^2}{2l^2}),
\end{equation} 
with the length scale~$l$ and the error variance $\sigma^2$ was chosen. These hyperparameters are optimized with the mean function, maximizing the log marginal likelihood and minimizing the cross-validation loss.
The mean function is often used to incorporate prior knowledge and is commonly set to zero if no approximation model is known.

The input~$\mathbf{x}$ is defined for three different configurations:
\begin{equation}
    \mathbf{x} =
    \begin{cases}
        (\theta_{J},\omega_{J})^T & \text{\small baseline}\\
        (\theta_{J},\omega_{J},\mathbf{M}_\text{EMG})^T & \text{\small EMG}\\
        (\theta_{J},\omega_{J},\mathbf{M}_\text{FMG})^T & \text{\small FMG}\\
    \end{cases}
\end{equation}
where $\theta_{J}$ represents the joint angle, $\omega_{J}$ the joint angular velocity, and $\mathbf{M}$ the muscle signals obtained either from \gls{FMG} or \gls{EMG} signals. These result in the corresponding estimated joint torques $\tilde T_{J,\text{baseline}}$, $\tilde T_{J,\text{FMG}}$ and $\tilde T_{J,\text{EMG}}$.

For each joint, the muscle signal $\mathbf{M}$ was selected based on the primary muscles involved in the motion of that joint~\cite{neumann_kinesiology_2002}:
\begin{equation}
    \mathbf{M} =
    \begin{cases}
        (M_\text{TA}, M_\text{GM},M_\text{GL}) & \text{\small ankle joint}\\
        (M_\text{BF},M_\text{RF},M_\text{ST},M_\text{VM},M_\text{VL}) & \text{\small knee joint}
    \end{cases}
    \label{eq:MuscleSignals}
\end{equation}
where $M_\text{TA}$, $M_\text{GM}$, $M_\text{GL}$, $M_\text{BF}$, $M_\text{RF}$, $M_\text{ST}$, $M_\text{VM}$ and $M_\text{VL}$ correspond to the muscular signals of the \emph{\gls{TA}}, \emph{\gls{GM}}, \emph{\gls{GL}}, \emph{\gls{BF}}, \emph{\gls{RF}}, \emph{\gls{ST}}, \emph{\gls{VM}} and \emph{\gls{VL}}.
The effects of biarticular muscles, which influence the movements of multiple joints at the same time, have not been considered beyond the joints described in \cref{eq:MuscleSignals}.

\subsection{User Study}

To validate the model combining joint biomechanics and muscle signals in isokinetic motion, a user study was carried out in a controlled laboratory environment using an IsoMed 2000 device. 
This device includes an integrated mode that facilitates isokinetic movement of a single joint at a time (\cref{fig:ExperimentalSetup}) and offers a direct on-axis measurement of the true sagittal joint torque. 
During isokinetic movement, the velocity of limb motion remains constant while the muscle forces can change. 
This provides a controlled experimental setup with minimal disturbances or external forces acting on the sensors during the motion.

In the study, ten healthy, able-bodied adults participated ($\text{m}=5\mid\text{f}=5$, age $26.8 \pm 3.2$ years, height $175.2 \pm 6.78$ \si{\cm}, weight $65.0 \pm 6.8$ \si{\kg}). 
%
%
The experimental protocol was approved by the Karlsruhe Institute of Technology (KIT) Ethics Committee under the JuBot project. 
Participants provided their informed consent in writing prior to the experiment, and all methods were performed following the Declaration of Helsinki.

The experiment was conducted with participants positioned on the IsoMed device, where their left leg was secured on either a foot or shank support. 
This arrangement permitted pure sagittal movement of the ankle or knee joint as shown in \cref{fig:ExperimentalSetup}. 
Participants were given up to \SI{10}{\min} to familiarize themselves with the IsoMed device, during which the device's rotation axis was carefully aligned with the sagittal axis of rotation of the joint using a built-in laser pointer. 
Furthermore, the mechanical end stops of the device were positioned to correspond to the user's maximum \gls{RoM} within their anatomical limits.
Following this, the participants performed two tasks in a random order:
\begin{figure}[b]
    \centering
    \hfill
    \begin{subfigure}[b]{0.45\columnwidth}
        \centering
        \includegraphics[width=\columnwidth]{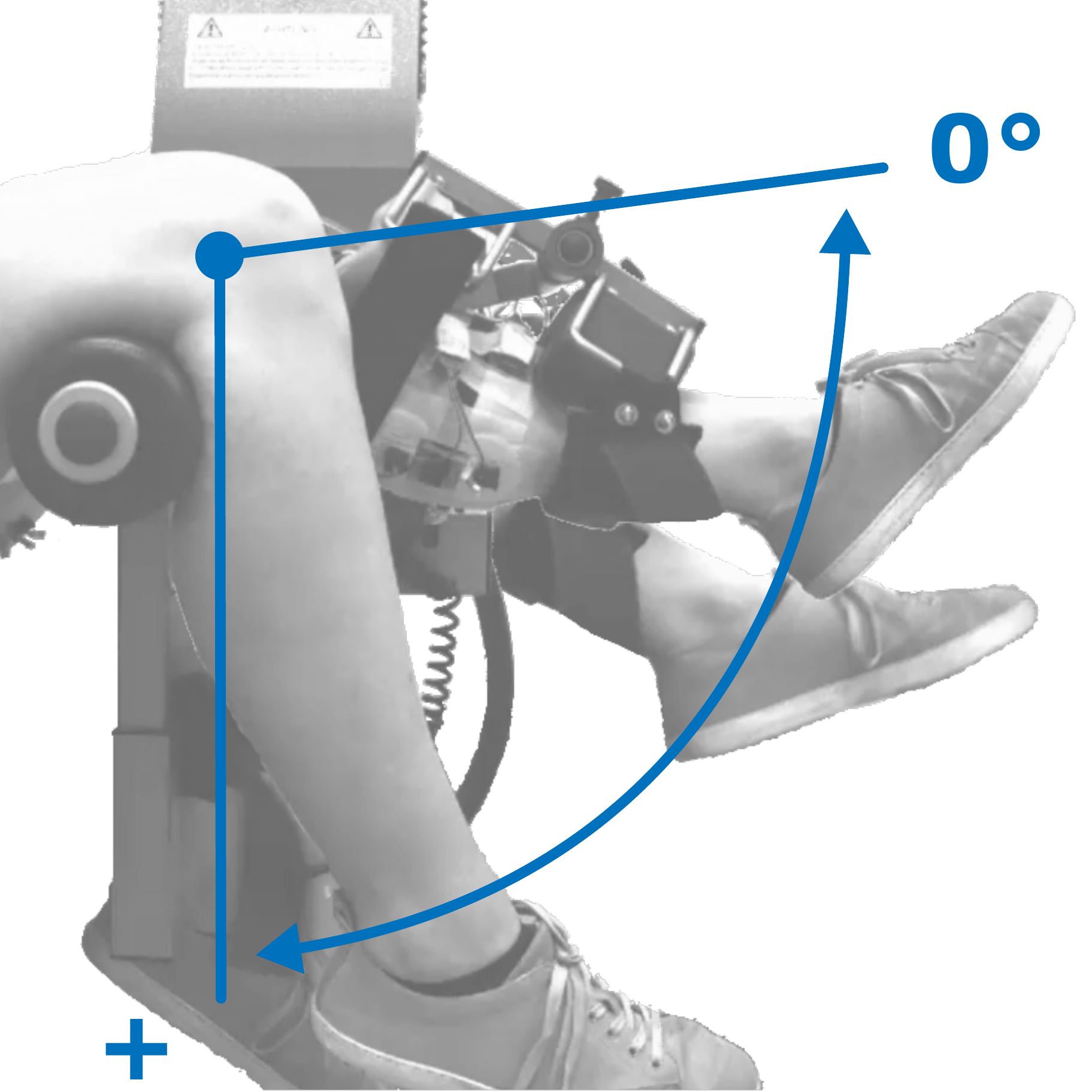}
        \caption{}
        \label{subfig:KneeSetup}
    \end{subfigure}
    \hfill
    \begin{subfigure}[b]{0.45\columnwidth}
        \centering
        \includegraphics[width=\columnwidth]{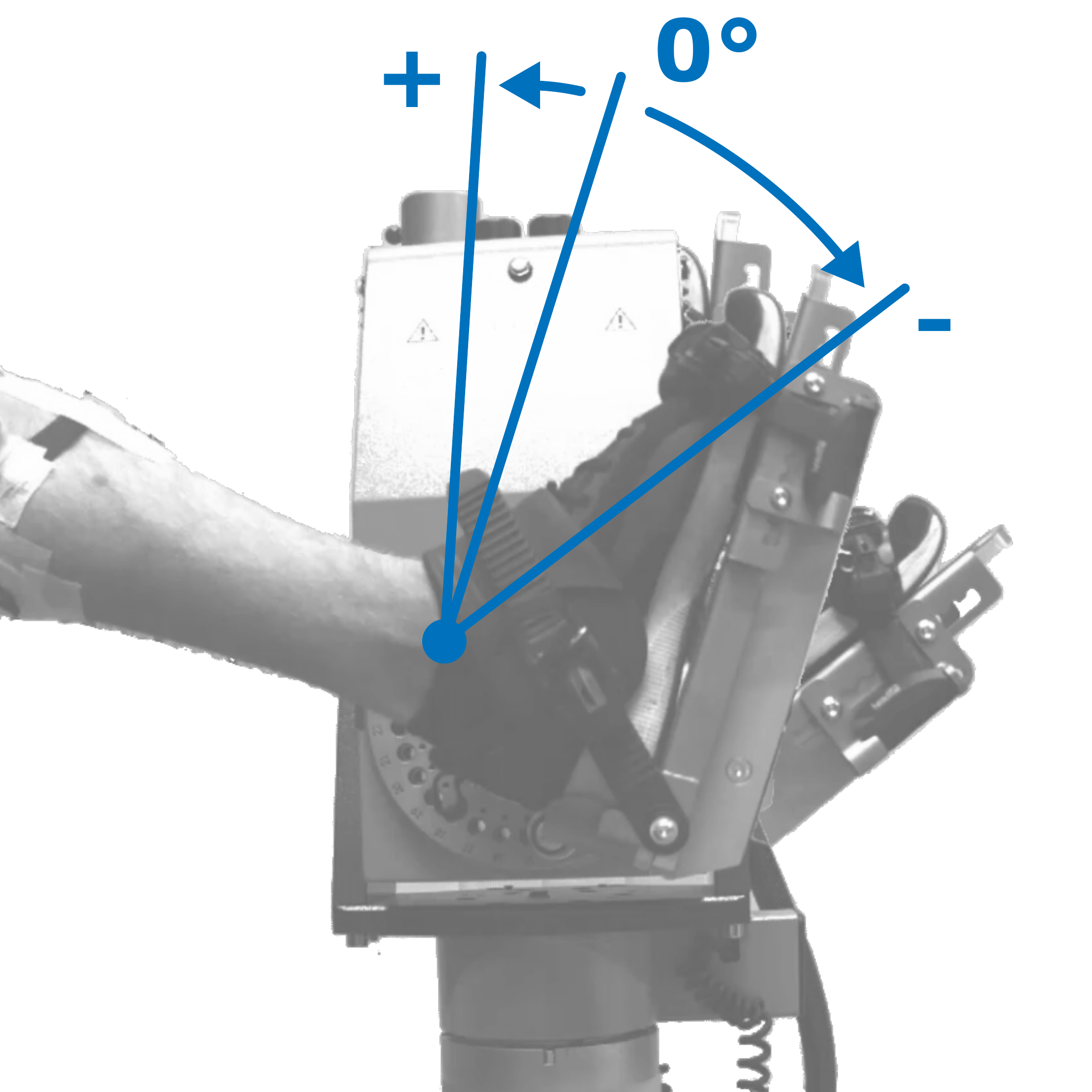}
        \caption{}
        \label{subfig:AnkleSetup}
    \end{subfigure}
    \hfill\null
        \caption{participant set-up on the IsoMed system for knee motion (a) and ankle motion (b) and the corresponding definition of the direction of the joint angle~$\theta_{J}$.}
        \label{fig:ExperimentalSetup}
\end{figure}
\begin{itemize}
	\item \emph{Knee}: 
    First, the knee joint was positioned at the maximum extension for initialization (\cref{subfig:KneeSetup}).
    Next, the participant performed a series of five swing motions, including flexion and extension within their maximum active range of motion, maintaining a constant maximum angular velocity.
    These motions were carried out at four different angular velocities: \SI{60}{\dps}, \SI{90}{\dps}, \SI{120}{\dps}, and \SI{150}{\dps}. 
    The initialization procedure and the five swing motions were repeated three times for each velocity, resulting in three recordings per angular velocity.
	\item \emph{Ankle}:
    The ankle joint was first initialized in a position in which the foot was orthogonal to the shank (\cref{subfig:AnkleSetup}).
    Next, the participant performed five swing motions, including dorsi- and plantarflexion between the maximum angles of their active joint range.
    The initialization procedure and the five swing motions were repeated three times for velocities: \SI{30}{\dps}, \SI{60}{\dps}, \SI{90}{\dps} and \SI{120}{\dps}, resulting in three recordings per angular velocity.
\end{itemize}

For calibration purposes, the \gls{FMG} sensors were initialized before accessing the IsoMed device by standing upright and relaxed on both feet for approximately 10 seconds.
Calibration measurements of the joint angle were conducted at each initial position $\theta_{J} = \SI{0}{\degree}$ as marked in \cref{fig:ExperimentalSetup}.

\subsection{Sensor Setup}

\begin{figure}[hb]
     \centering
     \fontsize{8pt}{10pt}\selectfont
     \hfill
     \begin{subfigure}[b]{0.42\columnwidth}
         \centering
         \def\svgwidth{\columnwidth}
         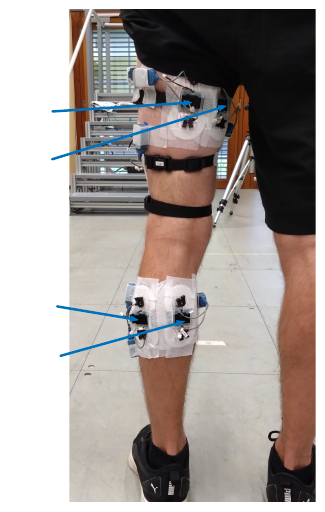
         \caption{}
         \label{subfig:SensorPositionsFront}
     \end{subfigure}
     \hfill
     \begin{subfigure}[b]{0.42\columnwidth}
         \centering
         \def\svgwidth{\columnwidth}
         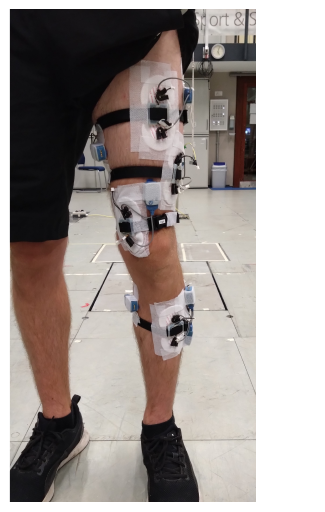
         \caption{}
         \label{subfig:SensorPositionsBack}
     \end{subfigure}
     \hfill\null
        \caption{\gls{EMG} (white electrodes) and \gls{FMG} (black straps) sensor positions on the back (a) and front (b) of the left leg.}
        \label{fig:SensorPosition}
\end{figure}

The used \gls{FMG} sensor unit measures the normal force resulting from a change in volume and stiffness of the human muscle underneath the cuff during leg motion. 
The sensor unit comprises five barometric pressure sensors on a single \gls{PCB}, covered by a silicon dome. 
Variations in pressure detected by these sensors reflect changes in the forces applied to the silicon dome. A detailed description of the sensor unit is given in~\cite{Marquardt2022},

In our experiments, eight \gls{FMG} units and eight \gls{EMG} electrode pairs were placed at anatomically relevant locations to measure the muscle activity of \emph{\gls{RF}}, \emph{\gls{BF}}, \emph{\gls{ST}}, \emph{\gls{VM}}, \emph{\gls{VL}}, \emph{\gls{GM}}, \emph{\gls{GL}} and \emph{\gls{TA}} as displayed in \cref{fig:SensorPosition}. 
The positions were determined based on \gls{EMG} placement recommendations given in  SENIAM~\cite{hermens_seniam_1999} combined with an assessment of real-time feedback of the \gls{EMG} sensor. 
The two \gls{EMG} electrodes were attached above and below the \gls{FMG} sensor unit along the course of the muscle to ensure measurement at the same point on the muscle.
In addition, the respective angular position~$\theta_{J}$ and torque of the joints~$T_{J}$ were recorded via the IsoMed device.

\subsection{Signal Processing}

The amplitudes of the \gls{EMG} signals are stochastic, and the signal fluctuates rapidly around zero. 
Therefore, the signals were band-pass filtered between \SIrange{20}{500}{\Hz}, rectified and afterwards low-pass filtered at \SI{6}{\Hz}. 
Both filters applied were fourth-order bi-directional Butterworth filters to achieve zero phase distortion.
The baseline offset was removed using the mean values obtained from the calibration measurements to calibrate both the \gls{FMG} signal and the joint angle signal. 
No further filtering was applied to the \gls{FMG} signal.

The angular joint velocity was derived based on the joint angle measurements. 
A second-order Butterworth filter with a cutoff frequency of \SI{20}{\Hz} was applied bi-directionally before calculating the gradient.
The segmentation of each complete motion, including flexion and extension of the knee joint and dorsi- and plantarflexion of the ankle joint, was performed based on the filtered joint angle to make it easier to recognize their extrema corresponding to the change in direction.

The \gls{FMG} sensor units allow a maximum sampling rate of \SI{200}{\hertz}, while the joint angle, joint torque, and \gls{EMG} signals were sampled at \SI{2000}{\hertz}.
To align and concatenate all data, each dataset was linearly interpolated to an equidistant number of data points, resulting in a down-sampling of the joint angle and \gls{EMG} data to fit the \gls{FMG} data.
To ensure that the data can be easily compared, all values were standardized using Z-score normalization, and after regression, the estimated joint torque was denormalized to its original scale.

To compare both the inter- and intra-participant performance of the model, the evaluation of the estimation results was based on a \gls{LOPO} and a \gls{LOVO} cross-validation.
The \gls{LOPO} method integrates the concatenated data of all four velocities for each participant, and the \gls{LOVO} is performed separately for each participant.
Since the computational complexity of \gls{GPR} is $O(n^3)$, a \gls{SGPR} was used to train the model for \gls{LOPO}. 
The initial inducing points are a randomly selected subset of \SI{0.25}{\percent} and \SI{10}{\percent} of the training data for \gls{LOPO} and \gls{LOVO}, respectively.
Early stopping was used with a patience of 50 epochs to avoid overfitting, tracking the validation loss based on \SI{20}{\percent} of the training data, and restoring the best model if no improvement was made over these 50 epochs. 

The standard deviation of the model estimation was evaluated by \gls{RMSE}.
A lower value of \gls{RMSE} implies a higher accuracy of the \gls{GPR} model.
Model training and testing were done using TensorFlow.

%% file: figures/sensor_positions_back.pdf_tex
\begingroup%
  \makeatletter%
  \providecommand\color[2][]{%
    \errmessage{(Inkscape) Color is used for the text in Inkscape, but the package 'color.sty' is not loaded}%
    \renewcommand\color[2][]{}%
  }%
  \providecommand\transparent[1]{%
    \errmessage{(Inkscape) Transparency is used (non-zero) for the text in Inkscape, but the package 'transparent.sty' is not loaded}%
    \renewcommand\transparent[1]{}%
  }%
  \providecommand\rotatebox[2]{#2}%
  \newcommand*\fsize{\dimexpr\f@size pt\relax}%
  \newcommand*\lineheight[1]{\fontsize{\fsize}{#1\fsize}\selectfont}%
  \ifx\svgwidth\undefined%
    \setlength{\unitlength}{93.54330709bp}%
    \ifx\svgscale\undefined%
      \relax%
    \else%
      \setlength{\unitlength}{\unitlength * \real{\svgscale}}%
    \fi%
  \else%
    \setlength{\unitlength}{\svgwidth}%
  \fi%
  \global\let\svgwidth\undefined%
  \global\let\svgscale\undefined%
  \makeatother%
  \begin{picture}(1,1.57575758)%
    \lineheight{1}%
    \setlength\tabcolsep{0pt}%
    \put(0.02888323,1.20925977){\makebox(0,0)[lt]{\smash{\begin{tabular}[t]{l}BF\end{tabular}}}}%
    \put(0.03163932,1.05689899){\makebox(0,0)[lt]{\smash{\begin{tabular}[t]{l}ST\end{tabular}}}}%
    \put(0.03243271,0.60235354){\makebox(0,0)[lt]{\smash{\begin{tabular}[t]{l}GL\end{tabular}}}}%
    \put(0.03243271,0.45083839){\makebox(0,0)[lt]{\smash{\begin{tabular}[t]{l}GM\end{tabular}}}}%
    \put(0,0){\includegraphics[width=\unitlength,page=1]{sensor_positions_back.pdf}}%
  \end{picture}%
\endgroup%

%% file: figures/sensor_positions_front.pdf_tex
\begingroup%
  \makeatletter%
  \providecommand\color[2][]{%
    \errmessage{(Inkscape) Color is used for the text in Inkscape, but the package 'color.sty' is not loaded}%
    \renewcommand\color[2][]{}%
  }%
  \providecommand\transparent[1]{%
    \errmessage{(Inkscape) Transparency is used (non-zero) for the text in Inkscape, but the package 'transparent.sty' is not loaded}%
    \renewcommand\transparent[1]{}%
  }%
  \providecommand\rotatebox[2]{#2}%
  \newcommand*\fsize{\dimexpr\f@size pt\relax}%
  \newcommand*\lineheight[1]{\fontsize{\fsize}{#1\fsize}\selectfont}%
  \ifx\svgwidth\undefined%
    \setlength{\unitlength}{93.54330709bp}%
    \ifx\svgscale\undefined%
      \relax%
    \else%
      \setlength{\unitlength}{\unitlength * \real{\svgscale}}%
    \fi%
  \else%
    \setlength{\unitlength}{\svgwidth}%
  \fi%
  \global\let\svgwidth\undefined%
  \global\let\svgscale\undefined%
  \makeatother%
  \begin{picture}(1,1.57575758)%
    \lineheight{1}%
    \setlength\tabcolsep{0pt}%
    \put(0,0){\includegraphics[width=\unitlength,page=1]{sensor_positions_front.pdf}}%
    \put(0.85616871,1.20925966){\makebox(0,0)[lt]{\smash{\begin{tabular}[t]{l}RF\end{tabular}}}}%
    \put(0.86343447,1.05774451){\makebox(0,0)[lt]{\smash{\begin{tabular}[t]{l}VL\end{tabular}}}}%
    \put(0.84739902,0.90622935){\makebox(0,0)[lt]{\smash{\begin{tabular}[t]{l}VM\end{tabular}}}}%
    \put(0.86452023,0.60319905){\makebox(0,0)[lt]{\smash{\begin{tabular}[t]{l}TA\end{tabular}}}}%
    \put(0,0){\includegraphics[width=\unitlength,page=2]{sensor_positions_front.pdf}}%
  \end{picture}%
\endgroup%

%% file: sections/results.tex
\section{RESULTS AND ANALYSIS}
\label{sec:results}

\begin{figure*}[ht]
    \vspace{1mm}
    \centering
    \begin{minipage}[c]{0.99\textwidth}
    \centering
    \fontsize{8pt}{10pt}\selectfont
    \begin{subfigure}[]{0.6\linewidth}
        \centering
        \def\svgwidth{\linewidth}
        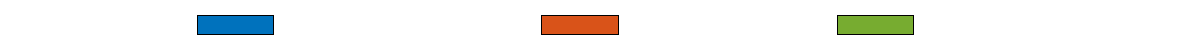
    \end{subfigure}
    \hfill
    \begin{subfigure}[]{0.49\linewidth}
        \centering
        \def\svgwidth{\linewidth}
        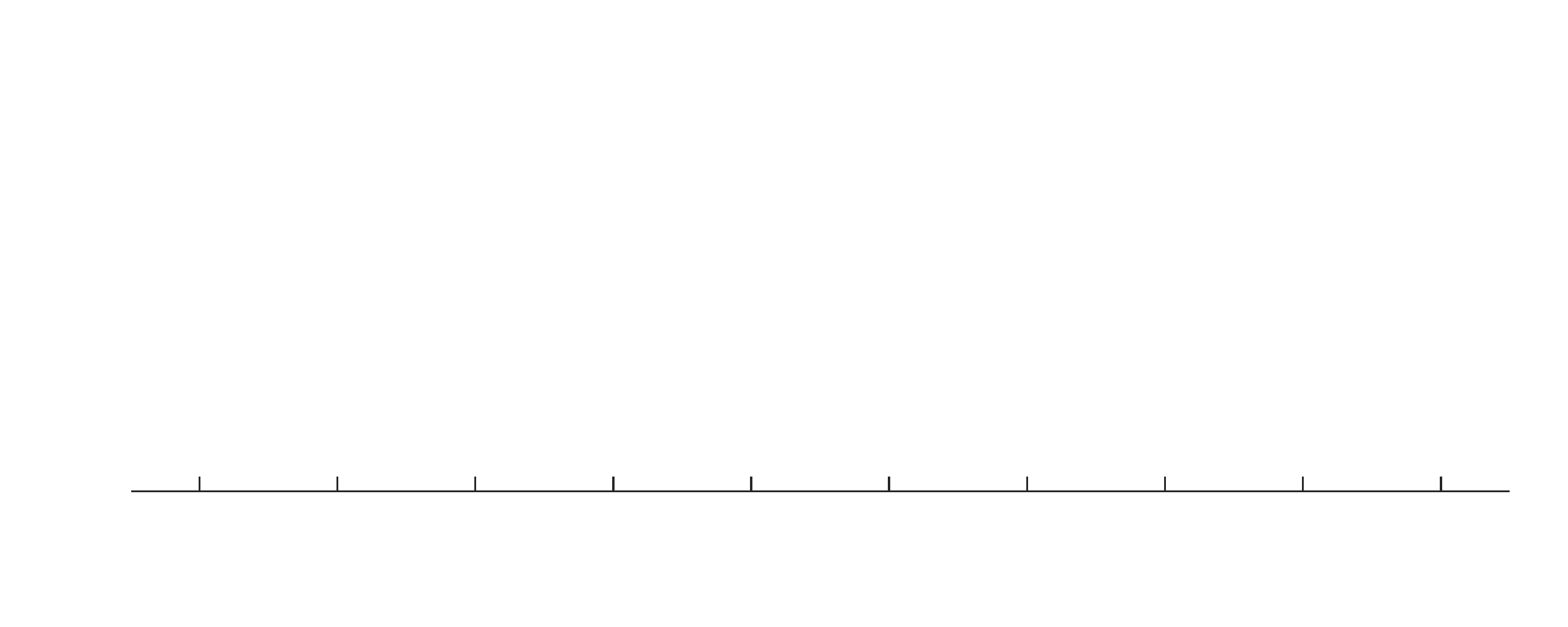
        \vspace{-2mm}
        \label{subfig:ResultsRMSEankleLOPO}
    \end{subfigure}
    \hfill
    \begin{subfigure}[]{0.49\linewidth}
        \centering
        \def\svgwidth{\linewidth}
        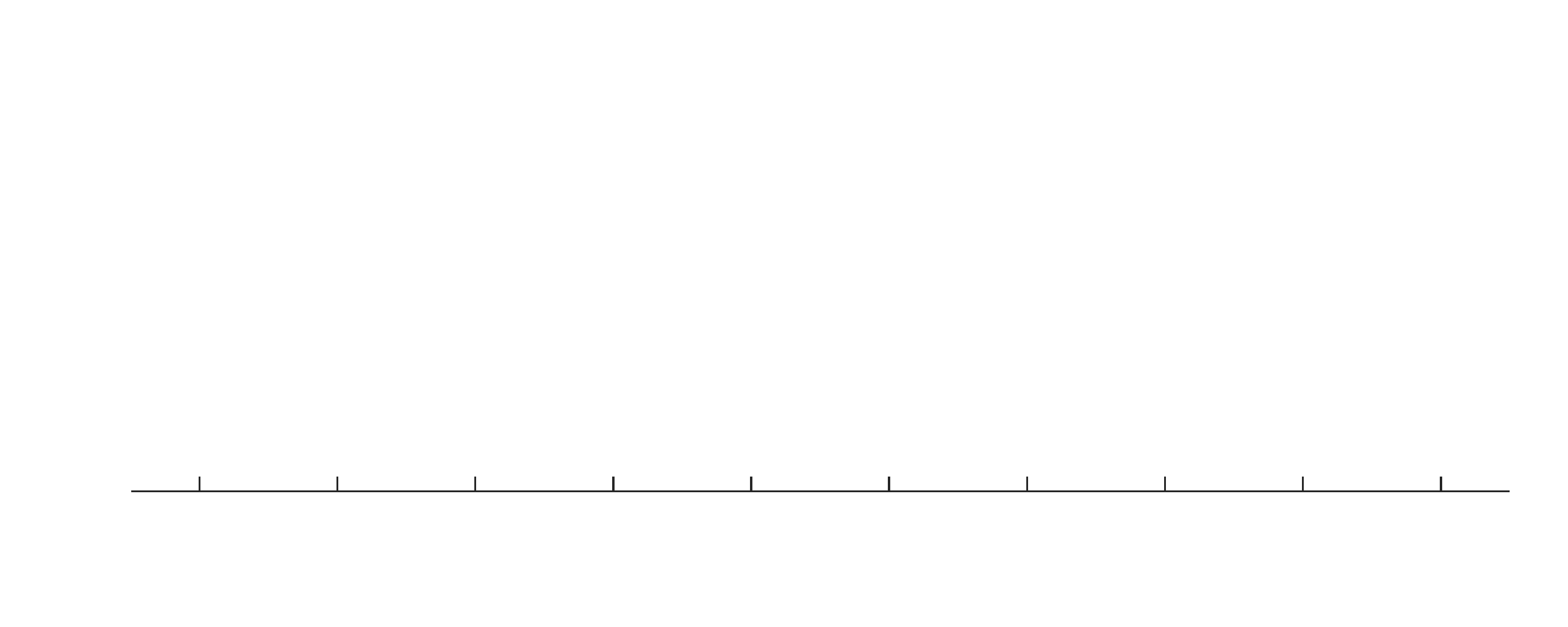
        \vspace{-2mm}
        \label{subfig:ResultsRMSEankleLOVO}
    \end{subfigure}
    \hfill
    \begin{subfigure}[]{0.49\linewidth}
        \centering
        \def\svgwidth{\linewidth}
        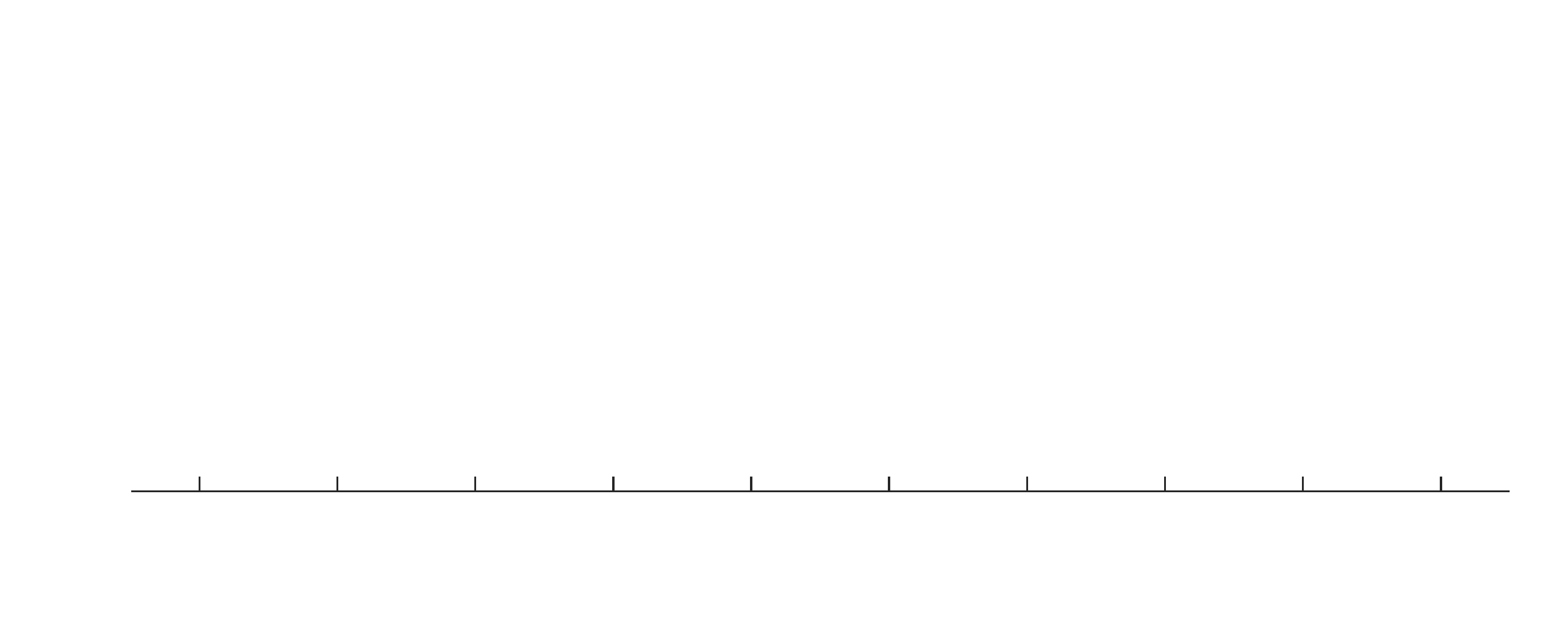
        \caption{\gls{LOPO}}
        \label{subfig:ResultsRMSEkneeLOPO}
    \end{subfigure}    
    \hfill
    \begin{subfigure}[]{0.49\linewidth}
        \centering
        \def\svgwidth{\linewidth}
        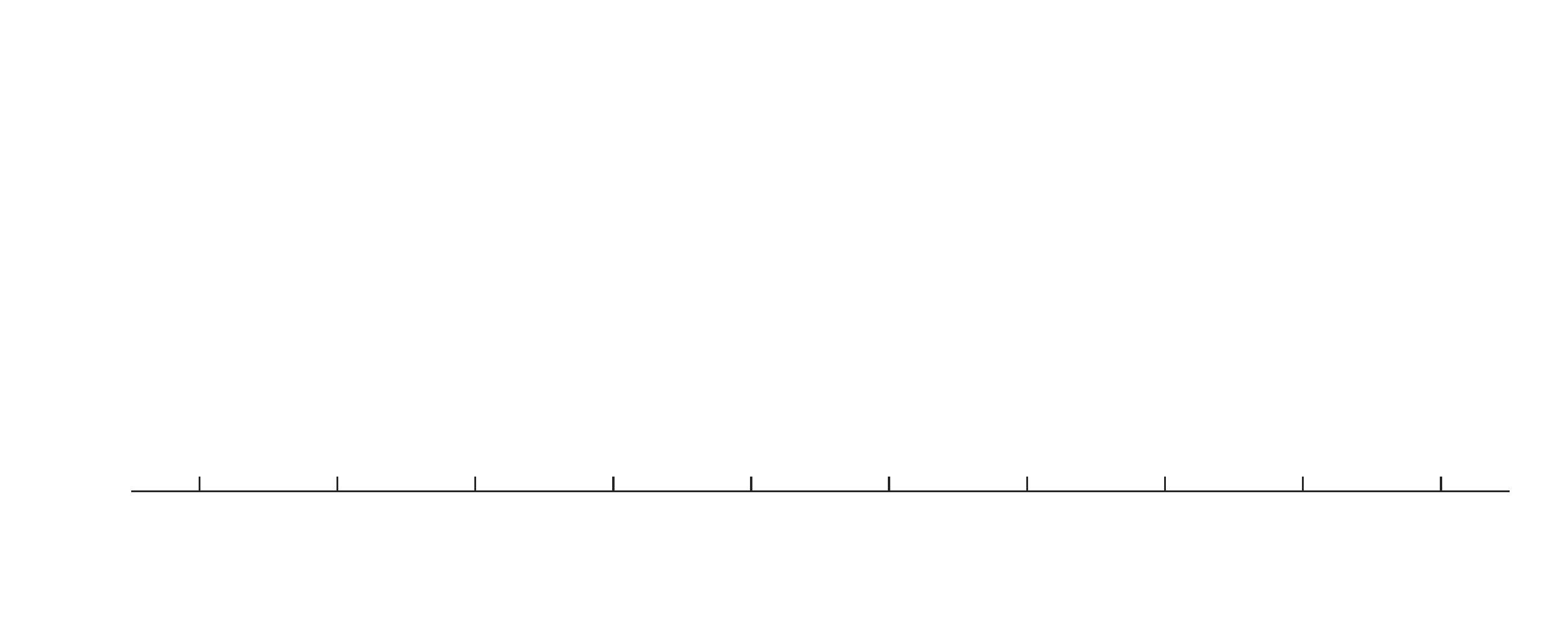
        \caption{\gls{LOVO}}
        \label{subfig:ResultsRMSEkneeLOVO}
    \end{subfigure}
    \end{minipage}
    \caption{\gls{RMSE} of the \gls{LOPO} (a) and mean and standard deviation of the \gls{RMSE} over all four velocities of the \gls{LOVO} (b) validation for all three model configurations, baseline (blue, left), \gls{EMG} (orange, middle) and \gls{FMG} (green, right) of both the ankle (top) and knee (bottom) joint.}
	\label{fig:ResultsValidationMetrics}
\end{figure*}

The proposed \gls{GPR}-based joint torque estimation approach was evaluated using \gls{LOPO} and \gls{LOVO} cross-validations to assess its performance.
The \gls{RMSE} values from these validations are presented in \cref{fig:ResultsValidationMetrics} for both the ankle and knee joints and the three model configurations.
%
%
The performance of the approaches varies between the \gls{LOPO} and \gls{LOVO} validations. 
In the \gls{LOPO} validation, the \gls{RMSE} values indicate that the baseline model is the most accurate of the three approaches. Conversely, the \gls{LOVO} validation shows that the extended model with \gls{FMG} signals achieves comparable or better performance than the baseline model.
In the \gls{LOPO} validation, the \gls{EMG}-based model outperforms the \gls{FMG}-based model for most participants but performs worse in the \gls{LOVO} validation.
However, the \gls{LOPO} results indicate that the \gls{FMG}-based model performs better for the knee joint than for the ankle joint, outperforming the \gls{EMG}-based model for four participants.

\begin{figure*}[hb]
    \centering
    \begin{minipage}[c]{0.99\textwidth}
    \centering
    \fontsize{8pt}{10pt}\selectfont
    \begin{subfigure}[]{0.6\linewidth}
        \centering
        \def\svgwidth{\linewidth}
        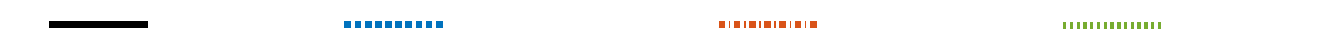
        \vspace{0.1mm}
    \end{subfigure}
    \hfill
    \begin{subfigure}[]{0.48\linewidth}
        \centering
        \def\svgwidth{\linewidth}
        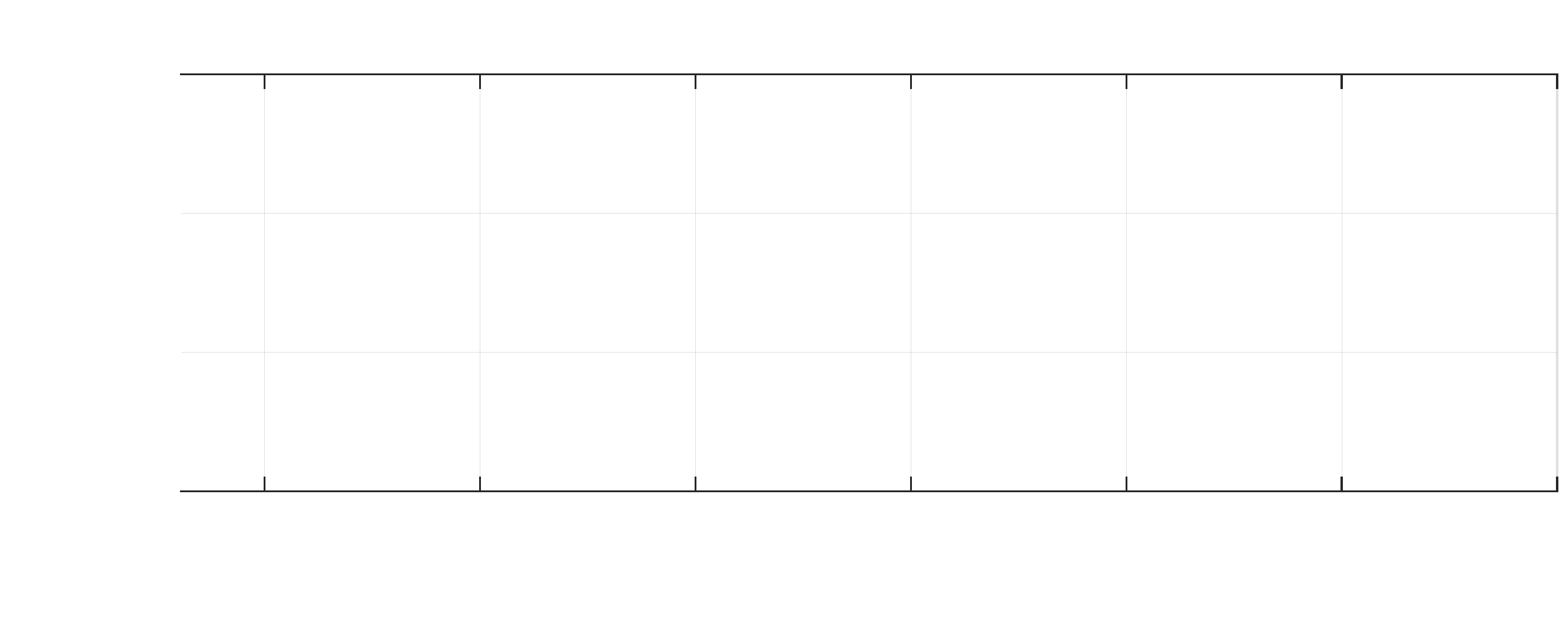
        \vspace{-2mm}
        \label{subfig:ResultsAnkleEstimationLOPO}
    \end{subfigure}
     \hfill
    \begin{subfigure}[]{0.48\linewidth}
        \centering
        \def\svgwidth{\linewidth}
        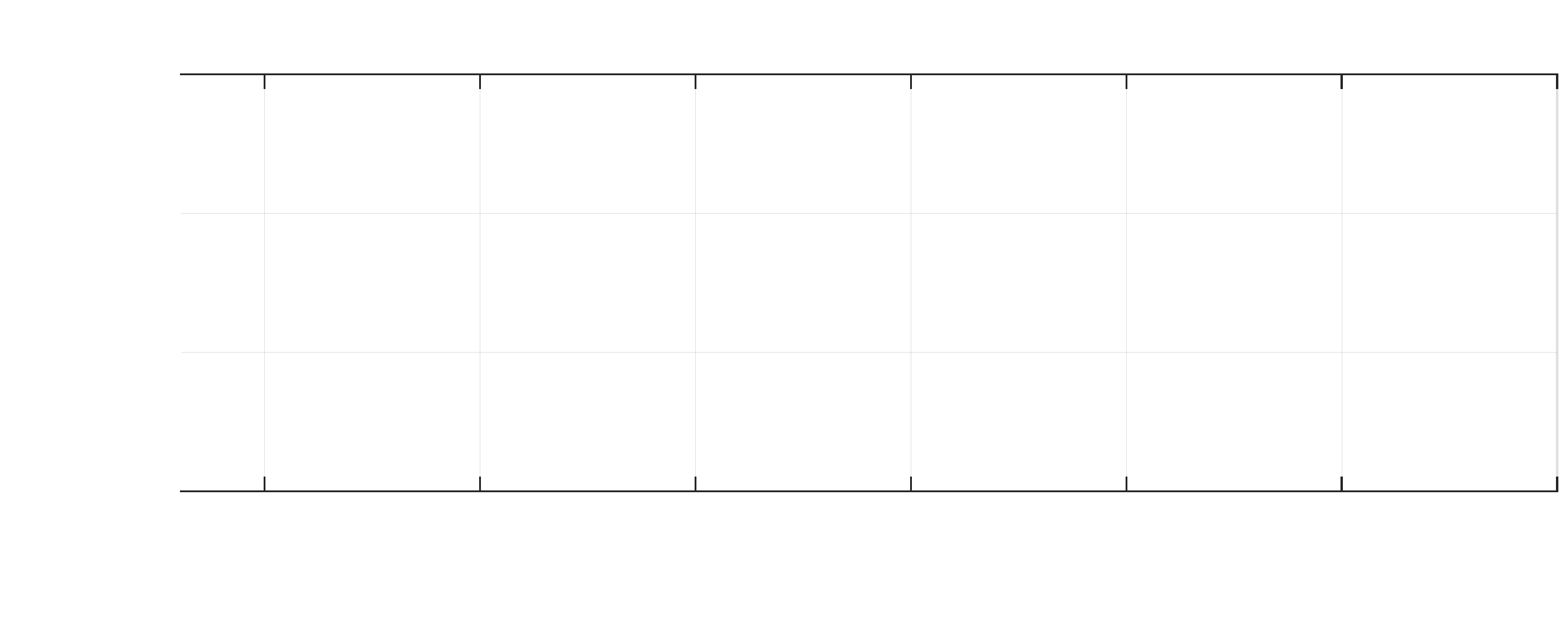
        \vspace{-2mm}
        \label{subfig:ResultsAnkleEstimationLOVO}
    \end{subfigure}
    \hfill
    \begin{subfigure}[]{0.48\linewidth}
        \centering
        \def\svgwidth{\linewidth}
        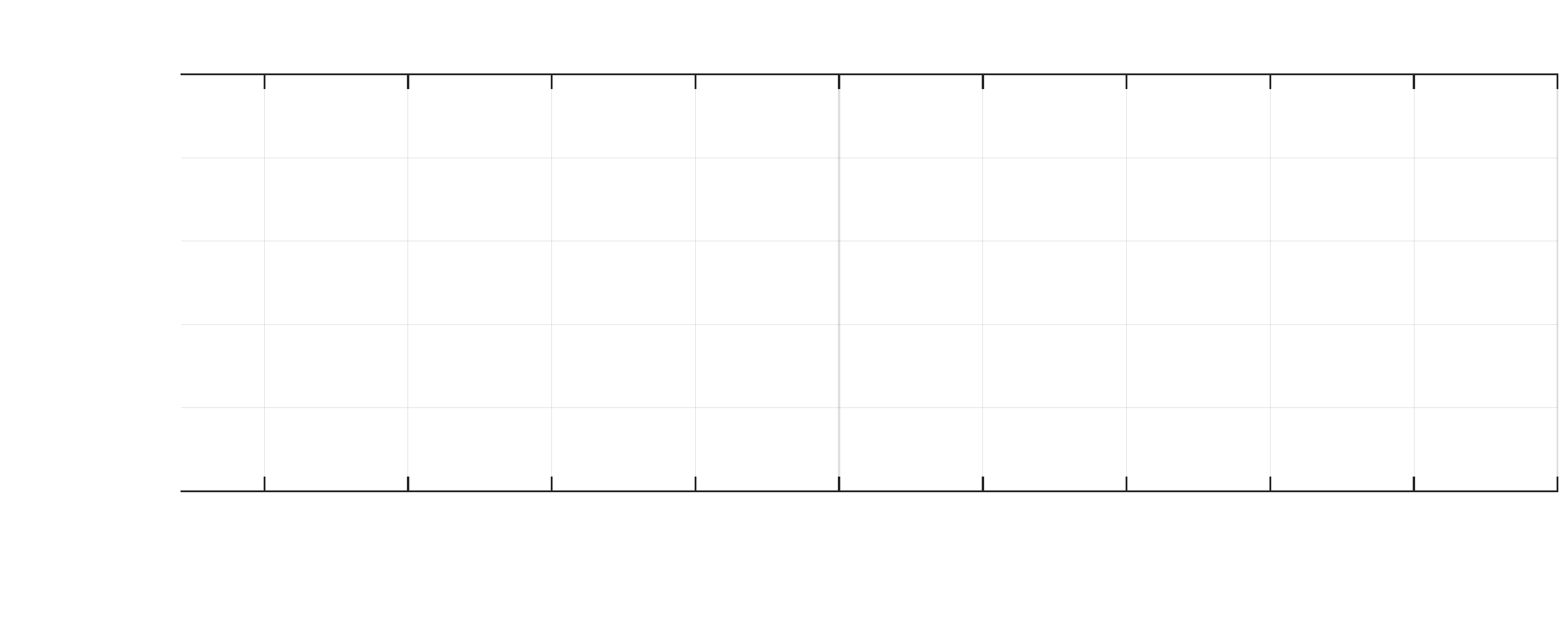
        \caption{LOPO}
        \label{subfig:ResultsKneeEstimationLOPO}
    \end{subfigure}
    \hfill
    \begin{subfigure}[]{0.48\linewidth}
        \centering
        \def\svgwidth{\linewidth}
        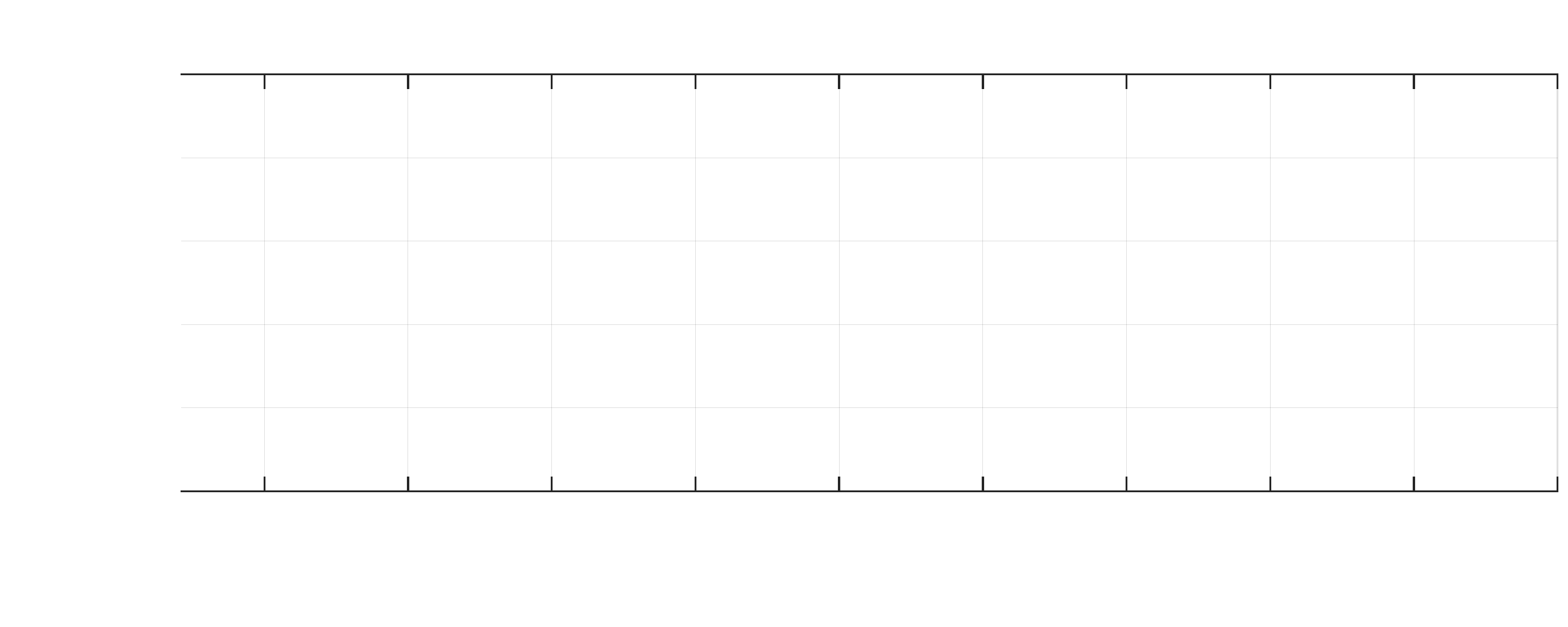
        \caption{LOVO}
        \label{subfig:ResultsKneeEstimationLOVO}
    \end{subfigure}
    \hfill
    \end{minipage}
    \caption{Measured and estimated torque over time for all three model configurations, baseline (blue, dashed), \gls{EMG} (orange, dash-dotted) and \gls{FMG} (green, dotted) of both the ankle (top) and knee (bottom) joint. The exemplary data of (a) the \gls{LOPO} and (b) the \gls{LOVO} validation is based on the testing data of one recording of the isokinetic motion of participant G at \SI{60}{\degree\per\second}.}
	\label{fig:ResultsEstimation}
\end{figure*}

\Cref{fig:ResultsEstimation} presents an example of joint torque estimation from time series data of a single participant, comparing estimated and measured joint torque during isokinetic motion at \SI{60}{\degree\per\second} based on the corresponding test data.
The baseline model, which considers only joint angle and velocity, provides a stable estimate but fails to capture the variations in the true torque amplitude. 
In contrast, incorporating muscle signals improves the model's ability to capture the variations in true torque amplitude that vary across participants and speeds.
However, this also results in greater variability compared to true torque amplitude in the \gls{EMG} or \gls{FMG} based torque estimates for \gls{LOVO} and \gls{LOPO}, respectively.

%% file: figures/results_RMSE_legend.pdf_tex
\begingroup%
  \makeatletter%
  \providecommand\color[2][]{%
    \errmessage{(Inkscape) Color is used for the text in Inkscape, but the package 'color.sty' is not loaded}%
    \renewcommand\color[2][]{}%
  }%
  \providecommand\transparent[1]{%
    \errmessage{(Inkscape) Transparency is used (non-zero) for the text in Inkscape, but the package 'transparent.sty' is not loaded}%
    \renewcommand\transparent[1]{}%
  }%
  \providecommand\rotatebox[2]{#2}%
  \newcommand*\fsize{\dimexpr\f@size pt\relax}%
  \newcommand*\lineheight[1]{\fontsize{\fsize}{#1\fsize}\selectfont}%
  \ifx\svgwidth\undefined%
    \setlength{\unitlength}{340.15748031bp}%
    \ifx\svgscale\undefined%
      \relax%
    \else%
      \setlength{\unitlength}{\unitlength * \real{\svgscale}}%
    \fi%
  \else%
    \setlength{\unitlength}{\svgwidth}%
  \fi%
  \global\let\svgwidth\undefined%
  \global\let\svgscale\undefined%
  \makeatother%
  \begin{picture}(1,0.04166667)%
    \lineheight{1}%
    \setlength\tabcolsep{0pt}%
    \put(0.25284137,0.01122358){\makebox(0,0)[lt]{\lineheight{1.25}\smash{\begin{tabular}[t]{l}baseline\end{tabular}}}}%
    \put(0.5408121,0.01122358){\makebox(0,0)[lt]{\lineheight{1.25}\smash{\begin{tabular}[t]{l}EMG\end{tabular}}}}%
    \put(0.78875259,0.01122358){\makebox(0,0)[lt]{\lineheight{1.25}\smash{\begin{tabular}[t]{l}FMG\end{tabular}}}}%
    \put(0,0){\includegraphics[width=\unitlength,page=1]{results_RMSE_legend.pdf}}%
  \end{picture}%
\endgroup%

%% file: figures/plot_InterSubject_RMSE_ankle.pdf_tex
\begingroup%
  \makeatletter%
  \providecommand\color[2][]{%
    \errmessage{(Inkscape) Color is used for the text in Inkscape, but the package 'color.sty' is not loaded}%
    \renewcommand\color[2][]{}%
  }%
  \providecommand\transparent[1]{%
    \errmessage{(Inkscape) Transparency is used (non-zero) for the text in Inkscape, but the package 'transparent.sty' is not loaded}%
    \renewcommand\transparent[1]{}%
  }%
  \providecommand\rotatebox[2]{#2}%
  \newcommand*\fsize{\dimexpr\f@size pt\relax}%
  \newcommand*\lineheight[1]{\fontsize{\fsize}{#1\fsize}\selectfont}%
  \ifx\svgwidth\undefined%
    \setlength{\unitlength}{750bp}%
    \ifx\svgscale\undefined%
      \relax%
    \else%
      \setlength{\unitlength}{\unitlength * \real{\svgscale}}%
    \fi%
  \else%
    \setlength{\unitlength}{\svgwidth}%
  \fi%
  \global\let\svgwidth\undefined%
  \global\let\svgscale\undefined%
  \makeatother%
  \begin{picture}(1,0.4)%
    \lineheight{1}%
    \setlength\tabcolsep{0pt}%
    \put(0,0){\includegraphics[width=\unitlength,page=1]{plot_InterSubject_RMSE_ankle.pdf}}%
    \put(0.12064019,0.05409355){\makebox(0,0)[lt]{\lineheight{1.25}\smash{\begin{tabular}[t]{l}A\end{tabular}}}}%
    \put(0.20863569,0.05409355){\makebox(0,0)[lt]{\lineheight{1.25}\smash{\begin{tabular}[t]{l}B\end{tabular}}}}%
    \put(0.29663119,0.05409355){\makebox(0,0)[lt]{\lineheight{1.25}\smash{\begin{tabular}[t]{l}C\end{tabular}}}}%
    \put(0.38462679,0.05409355){\makebox(0,0)[lt]{\lineheight{1.25}\smash{\begin{tabular}[t]{l}D\end{tabular}}}}%
    \put(0.47262229,0.05409355){\makebox(0,0)[lt]{\lineheight{1.25}\smash{\begin{tabular}[t]{l}E\end{tabular}}}}%
    \put(0.56061789,0.05409355){\makebox(0,0)[lt]{\lineheight{1.25}\smash{\begin{tabular}[t]{l}F\end{tabular}}}}%
    \put(0.64861339,0.05409355){\makebox(0,0)[lt]{\lineheight{1.25}\smash{\begin{tabular}[t]{l}G\end{tabular}}}}%
    \put(0.73660899,0.05409355){\makebox(0,0)[lt]{\lineheight{1.25}\smash{\begin{tabular}[t]{l}H\end{tabular}}}}%
    \put(0.82460449,0.05409355){\makebox(0,0)[lt]{\lineheight{1.25}\smash{\begin{tabular}[t]{l}I\end{tabular}}}}%
    \put(0.91260009,0.05409355){\makebox(0,0)[lt]{\lineheight{1.25}\smash{\begin{tabular}[t]{l}J\end{tabular}}}}%
    \put(0.45424023,0.02076025){\makebox(0,0)[lt]{\lineheight{1.25}\smash{\begin{tabular}[t]{l}Test Participant\end{tabular}}}}%
    \put(0,0){\includegraphics[width=\unitlength,page=2]{plot_InterSubject_RMSE_ankle.pdf}}%
    \put(0.05839453,0.07706025){\makebox(0,0)[lt]{\lineheight{1.25}\smash{\begin{tabular}[t]{l}0\end{tabular}}}}%
    \put(0.04304297,0.30506025){\makebox(0,0)[lt]{\lineheight{1.25}\smash{\begin{tabular}[t]{l}30\end{tabular}}}}%
    \put(0.03844143,0.09187543){\rotatebox{90}{\makebox(0,0)[lt]{\lineheight{1.25}\smash{\begin{tabular}[t]{l}RMSE (in \si{Nm})\end{tabular}}}}}%
    \put(0.47133235,0.36860361){\makebox(0,0)[lt]{\lineheight{1.25}\smash{\begin{tabular}[t]{l}\textbf{ankle}\end{tabular}}}}%
    \put(0,0){\includegraphics[width=\unitlength,page=3]{plot_InterSubject_RMSE_ankle.pdf}}%
  \end{picture}%
\endgroup%

%% file: figures/plot_IntraSubject_RMSE_ankle.pdf_tex
\begingroup%
  \makeatletter%
  \providecommand\color[2][]{%
    \errmessage{(Inkscape) Color is used for the text in Inkscape, but the package 'color.sty' is not loaded}%
    \renewcommand\color[2][]{}%
  }%
  \providecommand\transparent[1]{%
    \errmessage{(Inkscape) Transparency is used (non-zero) for the text in Inkscape, but the package 'transparent.sty' is not loaded}%
    \renewcommand\transparent[1]{}%
  }%
  \providecommand\rotatebox[2]{#2}%
  \newcommand*\fsize{\dimexpr\f@size pt\relax}%
  \newcommand*\lineheight[1]{\fontsize{\fsize}{#1\fsize}\selectfont}%
  \ifx\svgwidth\undefined%
    \setlength{\unitlength}{750bp}%
    \ifx\svgscale\undefined%
      \relax%
    \else%
      \setlength{\unitlength}{\unitlength * \real{\svgscale}}%
    \fi%
  \else%
    \setlength{\unitlength}{\svgwidth}%
  \fi%
  \global\let\svgwidth\undefined%
  \global\let\svgscale\undefined%
  \makeatother%
  \begin{picture}(1,0.4)%
    \lineheight{1}%
    \setlength\tabcolsep{0pt}%
    \put(0,0){\includegraphics[width=\unitlength,page=1]{plot_IntraSubject_RMSE_ankle.pdf}}%
    \put(0.45424023,0.02076025){\makebox(0,0)[lt]{\lineheight{1.25}\smash{\begin{tabular}[t]{l}Participant\end{tabular}}}}%
    \put(0,0){\includegraphics[width=\unitlength,page=2]{plot_IntraSubject_RMSE_ankle.pdf}}%
    \put(0.05839453,0.07706025){\makebox(0,0)[lt]{\lineheight{1.25}\smash{\begin{tabular}[t]{l}0\end{tabular}}}}%
    \put(0.04304202,0.30506025){\makebox(0,0)[lt]{\lineheight{1.25}\smash{\begin{tabular}[t]{l}30\end{tabular}}}}%
    \put(0.03843983,0.09187543){\rotatebox{90}{\makebox(0,0)[lt]{\lineheight{1.25}\smash{\begin{tabular}[t]{l}RMSE (in \si{Nm})\end{tabular}}}}}%
    \put(0.48924033,0.36860361){\makebox(0,0)[lt]{\lineheight{1.25}\smash{\begin{tabular}[t]{l}\textbf{ankle}\end{tabular}}}}%
    \put(0,0){\includegraphics[width=\unitlength,page=3]{plot_IntraSubject_RMSE_ankle.pdf}}%
    \put(0.12064019,0.05409201){\makebox(0,0)[lt]{\lineheight{1.25}\smash{\begin{tabular}[t]{l}A\end{tabular}}}}%
    \put(0.2086357,0.05409201){\makebox(0,0)[lt]{\lineheight{1.25}\smash{\begin{tabular}[t]{l}B\end{tabular}}}}%
    \put(0.2966312,0.05409201){\makebox(0,0)[lt]{\lineheight{1.25}\smash{\begin{tabular}[t]{l}C\end{tabular}}}}%
    \put(0.3846268,0.05409201){\makebox(0,0)[lt]{\lineheight{1.25}\smash{\begin{tabular}[t]{l}D\end{tabular}}}}%
    \put(0.4726223,0.05409201){\makebox(0,0)[lt]{\lineheight{1.25}\smash{\begin{tabular}[t]{l}E\end{tabular}}}}%
    \put(0.5606179,0.05409201){\makebox(0,0)[lt]{\lineheight{1.25}\smash{\begin{tabular}[t]{l}F\end{tabular}}}}%
    \put(0.6486134,0.05409201){\makebox(0,0)[lt]{\lineheight{1.25}\smash{\begin{tabular}[t]{l}G\end{tabular}}}}%
    \put(0.736609,0.05409201){\makebox(0,0)[lt]{\lineheight{1.25}\smash{\begin{tabular}[t]{l}H\end{tabular}}}}%
    \put(0.8246045,0.05409201){\makebox(0,0)[lt]{\lineheight{1.25}\smash{\begin{tabular}[t]{l}I\end{tabular}}}}%
    \put(0.9126001,0.05409201){\makebox(0,0)[lt]{\lineheight{1.25}\smash{\begin{tabular}[t]{l}J\end{tabular}}}}%
    \put(0,0){\includegraphics[width=\unitlength,page=4]{plot_IntraSubject_RMSE_ankle.pdf}}%
  \end{picture}%
\endgroup%

%% file: figures/plot_InterSubject_RMSE_knee.pdf_tex
\begingroup%
  \makeatletter%
  \providecommand\color[2][]{%
    \errmessage{(Inkscape) Color is used for the text in Inkscape, but the package 'color.sty' is not loaded}%
    \renewcommand\color[2][]{}%
  }%
  \providecommand\transparent[1]{%
    \errmessage{(Inkscape) Transparency is used (non-zero) for the text in Inkscape, but the package 'transparent.sty' is not loaded}%
    \renewcommand\transparent[1]{}%
  }%
  \providecommand\rotatebox[2]{#2}%
  \newcommand*\fsize{\dimexpr\f@size pt\relax}%
  \newcommand*\lineheight[1]{\fontsize{\fsize}{#1\fsize}\selectfont}%
  \ifx\svgwidth\undefined%
    \setlength{\unitlength}{750bp}%
    \ifx\svgscale\undefined%
      \relax%
    \else%
      \setlength{\unitlength}{\unitlength * \real{\svgscale}}%
    \fi%
  \else%
    \setlength{\unitlength}{\svgwidth}%
  \fi%
  \global\let\svgwidth\undefined%
  \global\let\svgscale\undefined%
  \makeatother%
  \begin{picture}(1,0.4)%
    \lineheight{1}%
    \setlength\tabcolsep{0pt}%
    \put(0,0){\includegraphics[width=\unitlength,page=1]{plot_InterSubject_RMSE_knee.pdf}}%
    \put(0.12064019,0.05409355){\makebox(0,0)[lt]{\lineheight{1.25}\smash{\begin{tabular}[t]{l}A\end{tabular}}}}%
    \put(0.20863569,0.05409355){\makebox(0,0)[lt]{\lineheight{1.25}\smash{\begin{tabular}[t]{l}B\end{tabular}}}}%
    \put(0.29663119,0.05409355){\makebox(0,0)[lt]{\lineheight{1.25}\smash{\begin{tabular}[t]{l}C\end{tabular}}}}%
    \put(0.38462679,0.05409355){\makebox(0,0)[lt]{\lineheight{1.25}\smash{\begin{tabular}[t]{l}D\end{tabular}}}}%
    \put(0.47262229,0.05409355){\makebox(0,0)[lt]{\lineheight{1.25}\smash{\begin{tabular}[t]{l}E\end{tabular}}}}%
    \put(0.56061789,0.05409355){\makebox(0,0)[lt]{\lineheight{1.25}\smash{\begin{tabular}[t]{l}F\end{tabular}}}}%
    \put(0.64861339,0.05409355){\makebox(0,0)[lt]{\lineheight{1.25}\smash{\begin{tabular}[t]{l}G\end{tabular}}}}%
    \put(0.73660899,0.05409355){\makebox(0,0)[lt]{\lineheight{1.25}\smash{\begin{tabular}[t]{l}H\end{tabular}}}}%
    \put(0.82460449,0.05409355){\makebox(0,0)[lt]{\lineheight{1.25}\smash{\begin{tabular}[t]{l}I\end{tabular}}}}%
    \put(0.91260009,0.05409355){\makebox(0,0)[lt]{\lineheight{1.25}\smash{\begin{tabular}[t]{l}J\end{tabular}}}}%
    \put(0.45424023,0.02076025){\makebox(0,0)[lt]{\lineheight{1.25}\smash{\begin{tabular}[t]{l}Test Participant\end{tabular}}}}%
    \put(0,0){\includegraphics[width=\unitlength,page=2]{plot_InterSubject_RMSE_knee.pdf}}%
    \put(0.05839453,0.07706025){\makebox(0,0)[lt]{\lineheight{1.25}\smash{\begin{tabular}[t]{l}0\end{tabular}}}}%
    \put(0.04304297,0.30506025){\makebox(0,0)[lt]{\lineheight{1.25}\smash{\begin{tabular}[t]{l}30\end{tabular}}}}%
    \put(0.03843983,0.09187543){\rotatebox{90}{\makebox(0,0)[lt]{\lineheight{1.25}\smash{\begin{tabular}[t]{l}RMSE (in \si{Nm})\end{tabular}}}}}%
    \put(0.49274033,0.36860361){\makebox(0,0)[lt]{\lineheight{1.25}\smash{\begin{tabular}[t]{l}\textbf{knee}\end{tabular}}}}%
    \put(0,0){\includegraphics[width=\unitlength,page=3]{plot_InterSubject_RMSE_knee.pdf}}%
  \end{picture}%
\endgroup%

%% file: figures/plot_IntraSubject_RMSE_knee.pdf_tex
\begingroup%
  \makeatletter%
  \providecommand\color[2][]{%
    \errmessage{(Inkscape) Color is used for the text in Inkscape, but the package 'color.sty' is not loaded}%
    \renewcommand\color[2][]{}%
  }%
  \providecommand\transparent[1]{%
    \errmessage{(Inkscape) Transparency is used (non-zero) for the text in Inkscape, but the package 'transparent.sty' is not loaded}%
    \renewcommand\transparent[1]{}%
  }%
  \providecommand\rotatebox[2]{#2}%
  \newcommand*\fsize{\dimexpr\f@size pt\relax}%
  \newcommand*\lineheight[1]{\fontsize{\fsize}{#1\fsize}\selectfont}%
  \ifx\svgwidth\undefined%
    \setlength{\unitlength}{750bp}%
    \ifx\svgscale\undefined%
      \relax%
    \else%
      \setlength{\unitlength}{\unitlength * \real{\svgscale}}%
    \fi%
  \else%
    \setlength{\unitlength}{\svgwidth}%
  \fi%
  \global\let\svgwidth\undefined%
  \global\let\svgscale\undefined%
  \makeatother%
  \begin{picture}(1,0.4)%
    \lineheight{1}%
    \setlength\tabcolsep{0pt}%
    \put(0,0){\includegraphics[width=\unitlength,page=1]{plot_IntraSubject_RMSE_knee.pdf}}%
    \put(0.45424023,0.02076025){\makebox(0,0)[lt]{\lineheight{1.25}\smash{\begin{tabular}[t]{l}Participant\end{tabular}}}}%
    \put(0,0){\includegraphics[width=\unitlength,page=2]{plot_IntraSubject_RMSE_knee.pdf}}%
    \put(0.05839453,0.07706025){\makebox(0,0)[lt]{\lineheight{1.25}\smash{\begin{tabular}[t]{l}0\end{tabular}}}}%
    \put(0.04304202,0.30506025){\makebox(0,0)[lt]{\lineheight{1.25}\smash{\begin{tabular}[t]{l}30\end{tabular}}}}%
    \put(0.03843983,0.09187543){\rotatebox{90}{\makebox(0,0)[lt]{\lineheight{1.25}\smash{\begin{tabular}[t]{l}RMSE (in \si{Nm})\end{tabular}}}}}%
    \put(0.47024954,0.36860361){\makebox(0,0)[lt]{\lineheight{1.25}\smash{\begin{tabular}[t]{l}\textbf{knee}\end{tabular}}}}%
    \put(0,0){\includegraphics[width=\unitlength,page=3]{plot_IntraSubject_RMSE_knee.pdf}}%
    \put(0.12064019,0.05409201){\makebox(0,0)[lt]{\lineheight{1.25}\smash{\begin{tabular}[t]{l}A\end{tabular}}}}%
    \put(0.20863569,0.05409201){\makebox(0,0)[lt]{\lineheight{1.25}\smash{\begin{tabular}[t]{l}B\end{tabular}}}}%
    \put(0.29663119,0.05409201){\makebox(0,0)[lt]{\lineheight{1.25}\smash{\begin{tabular}[t]{l}C\end{tabular}}}}%
    \put(0.38462679,0.05409201){\makebox(0,0)[lt]{\lineheight{1.25}\smash{\begin{tabular}[t]{l}D\end{tabular}}}}%
    \put(0.47262229,0.05409201){\makebox(0,0)[lt]{\lineheight{1.25}\smash{\begin{tabular}[t]{l}E\end{tabular}}}}%
    \put(0.56061789,0.05409201){\makebox(0,0)[lt]{\lineheight{1.25}\smash{\begin{tabular}[t]{l}F\end{tabular}}}}%
    \put(0.64861339,0.05409201){\makebox(0,0)[lt]{\lineheight{1.25}\smash{\begin{tabular}[t]{l}G\end{tabular}}}}%
    \put(0.73660899,0.05409201){\makebox(0,0)[lt]{\lineheight{1.25}\smash{\begin{tabular}[t]{l}H\end{tabular}}}}%
    \put(0.82460449,0.05409201){\makebox(0,0)[lt]{\lineheight{1.25}\smash{\begin{tabular}[t]{l}I\end{tabular}}}}%
    \put(0.91260009,0.05409201){\makebox(0,0)[lt]{\lineheight{1.25}\smash{\begin{tabular}[t]{l}J\end{tabular}}}}%
    \put(0,0){\includegraphics[width=\unitlength,page=4]{plot_IntraSubject_RMSE_knee.pdf}}%
  \end{picture}%
\endgroup%

%% file: figures/results_torqueEstimation_legend.pdf_tex
\begingroup%
  \makeatletter%
  \providecommand\color[2][]{%
    \errmessage{(Inkscape) Color is used for the text in Inkscape, but the package 'color.sty' is not loaded}%
    \renewcommand\color[2][]{}%
  }%
  \providecommand\transparent[1]{%
    \errmessage{(Inkscape) Transparency is used (non-zero) for the text in Inkscape, but the package 'transparent.sty' is not loaded}%
    \renewcommand\transparent[1]{}%
  }%
  \providecommand\rotatebox[2]{#2}%
  \newcommand*\fsize{\dimexpr\f@size pt\relax}%
  \newcommand*\lineheight[1]{\fontsize{\fsize}{#1\fsize}\selectfont}%
  \ifx\svgwidth\undefined%
    \setlength{\unitlength}{382.67716535bp}%
    \ifx\svgscale\undefined%
      \relax%
    \else%
      \setlength{\unitlength}{\unitlength * \real{\svgscale}}%
    \fi%
  \else%
    \setlength{\unitlength}{\svgwidth}%
  \fi%
  \global\let\svgwidth\undefined%
  \global\let\svgscale\undefined%
  \makeatother%
  \begin{picture}(1,0.03703704)%
    \lineheight{1}%
    \setlength\tabcolsep{0pt}%
    \put(0,0){\includegraphics[width=\unitlength,page=1]{results_torqueEstimation_legend.pdf}}%
    \put(0.35353192,0.00997652){\makebox(0,0)[lt]{\lineheight{1.25}\smash{\begin{tabular}[t]{l}$\tilde T_{J,\text{baseline}}$\end{tabular}}}}%
    \put(0.63770287,0.00997652){\makebox(0,0)[lt]{\lineheight{1.25}\smash{\begin{tabular}[t]{l}$\tilde T_{J,\text{EMG}}$\end{tabular}}}}%
    \put(0.89914115,0.00997652){\makebox(0,0)[lt]{\lineheight{1.25}\smash{\begin{tabular}[t]{l}$\tilde T_{J,\text{FMG}}$\end{tabular}}}}%
    \put(0.1419943,0.00997652){\makebox(0,0)[lt]{\lineheight{1.25}\smash{\begin{tabular}[t]{l}$T_{J}$\end{tabular}}}}%
  \end{picture}%
\endgroup%

%% file: figures/plot_InterSubject_torqueEstimation_ankle_p6547_s60.pdf_tex
\begingroup%
  \makeatletter%
  \providecommand\color[2][]{%
    \errmessage{(Inkscape) Color is used for the text in Inkscape, but the package 'color.sty' is not loaded}%
    \renewcommand\color[2][]{}%
  }%
  \providecommand\transparent[1]{%
    \errmessage{(Inkscape) Transparency is used (non-zero) for the text in Inkscape, but the package 'transparent.sty' is not loaded}%
    \renewcommand\transparent[1]{}%
  }%
  \providecommand\rotatebox[2]{#2}%
  \newcommand*\fsize{\dimexpr\f@size pt\relax}%
  \newcommand*\lineheight[1]{\fontsize{\fsize}{#1\fsize}\selectfont}%
  \ifx\svgwidth\undefined%
    \setlength{\unitlength}{750bp}%
    \ifx\svgscale\undefined%
      \relax%
    \else%
      \setlength{\unitlength}{\unitlength * \real{\svgscale}}%
    \fi%
  \else%
    \setlength{\unitlength}{\svgwidth}%
  \fi%
  \global\let\svgwidth\undefined%
  \global\let\svgscale\undefined%
  \makeatother%
  \begin{picture}(1,0.4)%
    \lineheight{1}%
    \setlength\tabcolsep{0pt}%
    \put(0,0){\includegraphics[width=\unitlength,page=1]{plot_InterSubject_torqueEstimation_ankle_p6547_s60.pdf}}%
    \put(0.16174796,0.05409355){\makebox(0,0)[lt]{\lineheight{1.25}\smash{\begin{tabular}[t]{l}0\end{tabular}}}}%
    \put(0.29920016,0.05409355){\makebox(0,0)[lt]{\lineheight{1.25}\smash{\begin{tabular}[t]{l}2\end{tabular}}}}%
    \put(0.43665226,0.05409355){\makebox(0,0)[lt]{\lineheight{1.25}\smash{\begin{tabular}[t]{l}4\end{tabular}}}}%
    \put(0.57410446,0.05409355){\makebox(0,0)[lt]{\lineheight{1.25}\smash{\begin{tabular}[t]{l}6\end{tabular}}}}%
    \put(0.71155656,0.05409355){\makebox(0,0)[lt]{\lineheight{1.25}\smash{\begin{tabular}[t]{l}8\end{tabular}}}}%
    \put(0.84250876,0.05409355){\makebox(0,0)[lt]{\lineheight{1.25}\smash{\begin{tabular}[t]{l}10\end{tabular}}}}%
    \put(0.97996086,0.05409355){\makebox(0,0)[lt]{\lineheight{1.25}\smash{\begin{tabular}[t]{l}12\end{tabular}}}}%
    \put(0.49590619,0.02088707){\makebox(0,0)[lt]{\lineheight{1.25}\smash{\begin{tabular}[t]{l}time (in \si{\s})\end{tabular}}}}%
    \put(0,0){\includegraphics[width=\unitlength,page=2]{plot_InterSubject_torqueEstimation_ankle_p6547_s60.pdf}}%
    \put(0.06499416,0.07706025){\makebox(0,0)[lt]{\lineheight{1.25}\smash{\begin{tabular}[t]{l}-50\end{tabular}}}}%
    \put(0.09399416,0.16572695){\makebox(0,0)[lt]{\lineheight{1.25}\smash{\begin{tabular}[t]{l}0\end{tabular}}}}%
    \put(0.07699416,0.25439355){\makebox(0,0)[lt]{\lineheight{1.25}\smash{\begin{tabular}[t]{l}50\end{tabular}}}}%
    \put(0.06099416,0.34306025){\makebox(0,0)[lt]{\lineheight{1.25}\smash{\begin{tabular}[t]{l}100\end{tabular}}}}%
    \put(0.03895546,0.1064547){\rotatebox{90}{\makebox(0,0)[lt]{\lineheight{1.25}\smash{\begin{tabular}[t]{l}torque (in \si{\Nm})\end{tabular}}}}}%
    \put(0.52583304,0.36860361){\makebox(0,0)[lt]{\lineheight{1.25}\smash{\begin{tabular}[t]{l}\textbf{ankle }\end{tabular}}}}%
    \put(0,0){\includegraphics[width=\unitlength,page=3]{plot_InterSubject_torqueEstimation_ankle_p6547_s60.pdf}}%
  \end{picture}%
\endgroup%

%% file: figures/plot_IntraSubject_torqueEstimation_ankle_p6547_s60.pdf_tex
\begingroup%
  \makeatletter%
  \providecommand\color[2][]{%
    \errmessage{(Inkscape) Color is used for the text in Inkscape, but the package 'color.sty' is not loaded}%
    \renewcommand\color[2][]{}%
  }%
  \providecommand\transparent[1]{%
    \errmessage{(Inkscape) Transparency is used (non-zero) for the text in Inkscape, but the package 'transparent.sty' is not loaded}%
    \renewcommand\transparent[1]{}%
  }%
  \providecommand\rotatebox[2]{#2}%
  \newcommand*\fsize{\dimexpr\f@size pt\relax}%
  \newcommand*\lineheight[1]{\fontsize{\fsize}{#1\fsize}\selectfont}%
  \ifx\svgwidth\undefined%
    \setlength{\unitlength}{750bp}%
    \ifx\svgscale\undefined%
      \relax%
    \else%
      \setlength{\unitlength}{\unitlength * \real{\svgscale}}%
    \fi%
  \else%
    \setlength{\unitlength}{\svgwidth}%
  \fi%
  \global\let\svgwidth\undefined%
  \global\let\svgscale\undefined%
  \makeatother%
  \begin{picture}(1,0.4)%
    \lineheight{1}%
    \setlength\tabcolsep{0pt}%
    \put(0,0){\includegraphics[width=\unitlength,page=1]{plot_IntraSubject_torqueEstimation_ankle_p6547_s60.pdf}}%
    \put(0.16174796,0.05409355){\makebox(0,0)[lt]{\lineheight{1.25}\smash{\begin{tabular}[t]{l}0\end{tabular}}}}%
    \put(0.29920016,0.05409355){\makebox(0,0)[lt]{\lineheight{1.25}\smash{\begin{tabular}[t]{l}2\end{tabular}}}}%
    \put(0.43665226,0.05409355){\makebox(0,0)[lt]{\lineheight{1.25}\smash{\begin{tabular}[t]{l}4\end{tabular}}}}%
    \put(0.57410446,0.05409355){\makebox(0,0)[lt]{\lineheight{1.25}\smash{\begin{tabular}[t]{l}6\end{tabular}}}}%
    \put(0.71155656,0.05409355){\makebox(0,0)[lt]{\lineheight{1.25}\smash{\begin{tabular}[t]{l}8\end{tabular}}}}%
    \put(0.84250876,0.05409355){\makebox(0,0)[lt]{\lineheight{1.25}\smash{\begin{tabular}[t]{l}10\end{tabular}}}}%
    \put(0.97996086,0.05409355){\makebox(0,0)[lt]{\lineheight{1.25}\smash{\begin{tabular}[t]{l}12\end{tabular}}}}%
    \put(0.49590619,0.02088707){\makebox(0,0)[lt]{\lineheight{1.25}\smash{\begin{tabular}[t]{l}time (in \si{\s})\end{tabular}}}}%
    \put(0,0){\includegraphics[width=\unitlength,page=2]{plot_IntraSubject_torqueEstimation_ankle_p6547_s60.pdf}}%
    \put(0.06499416,0.07706025){\makebox(0,0)[lt]{\lineheight{1.25}\smash{\begin{tabular}[t]{l}-50\end{tabular}}}}%
    \put(0.09399416,0.16572695){\makebox(0,0)[lt]{\lineheight{1.25}\smash{\begin{tabular}[t]{l}0\end{tabular}}}}%
    \put(0.07699416,0.25439355){\makebox(0,0)[lt]{\lineheight{1.25}\smash{\begin{tabular}[t]{l}50\end{tabular}}}}%
    \put(0.06099416,0.34306025){\makebox(0,0)[lt]{\lineheight{1.25}\smash{\begin{tabular}[t]{l}100\end{tabular}}}}%
    \put(0.03895546,0.1064547){\rotatebox{90}{\makebox(0,0)[lt]{\lineheight{1.25}\smash{\begin{tabular}[t]{l}torque (in \si{\Nm})\end{tabular}}}}}%
    \put(0.52583304,0.36860361){\makebox(0,0)[lt]{\lineheight{1.25}\smash{\begin{tabular}[t]{l}\textbf{ankle}\end{tabular}}}}%
    \put(0,0){\includegraphics[width=\unitlength,page=3]{plot_IntraSubject_torqueEstimation_ankle_p6547_s60.pdf}}%
  \end{picture}%
\endgroup%

%% file: figures/plot_InterSubject_torqueEstimation_knee_p6547_s60.pdf_tex
\begingroup%
  \makeatletter%
  \providecommand\color[2][]{%
    \errmessage{(Inkscape) Color is used for the text in Inkscape, but the package 'color.sty' is not loaded}%
    \renewcommand\color[2][]{}%
  }%
  \providecommand\transparent[1]{%
    \errmessage{(Inkscape) Transparency is used (non-zero) for the text in Inkscape, but the package 'transparent.sty' is not loaded}%
    \renewcommand\transparent[1]{}%
  }%
  \providecommand\rotatebox[2]{#2}%
  \newcommand*\fsize{\dimexpr\f@size pt\relax}%
  \newcommand*\lineheight[1]{\fontsize{\fsize}{#1\fsize}\selectfont}%
  \ifx\svgwidth\undefined%
    \setlength{\unitlength}{750bp}%
    \ifx\svgscale\undefined%
      \relax%
    \else%
      \setlength{\unitlength}{\unitlength * \real{\svgscale}}%
    \fi%
  \else%
    \setlength{\unitlength}{\svgwidth}%
  \fi%
  \global\let\svgwidth\undefined%
  \global\let\svgscale\undefined%
  \makeatother%
  \begin{picture}(1,0.4)%
    \lineheight{1}%
    \setlength\tabcolsep{0pt}%
    \put(0,0){\includegraphics[width=\unitlength,page=1]{plot_InterSubject_torqueEstimation_knee_p6547_s60.pdf}}%
    \put(0.16167025,0.05409355){\makebox(0,0)[lt]{\lineheight{1.25}\smash{\begin{tabular}[t]{l}0\end{tabular}}}}%
    \put(0.34500965,0.05409355){\makebox(0,0)[lt]{\lineheight{1.25}\smash{\begin{tabular}[t]{l}4\end{tabular}}}}%
    \put(0.52834895,0.05409355){\makebox(0,0)[lt]{\lineheight{1.25}\smash{\begin{tabular}[t]{l}8\end{tabular}}}}%
    \put(0.70518835,0.05409355){\makebox(0,0)[lt]{\lineheight{1.25}\smash{\begin{tabular}[t]{l}12\end{tabular}}}}%
    \put(0.88852765,0.05409355){\makebox(0,0)[lt]{\lineheight{1.25}\smash{\begin{tabular}[t]{l}16\end{tabular}}}}%
    \put(0.48814268,0.02088707){\makebox(0,0)[lt]{\lineheight{1.25}\smash{\begin{tabular}[t]{l}time (in \si{\s})\end{tabular}}}}%
    \put(0,0){\includegraphics[width=\unitlength,page=2]{plot_InterSubject_torqueEstimation_knee_p6547_s60.pdf}}%
    \put(0.04923065,0.13026025){\makebox(0,0)[lt]{\lineheight{1.25}\smash{\begin{tabular}[t]{l}-100\end{tabular}}}}%
    \put(0.09423065,0.23666025){\makebox(0,0)[lt]{\lineheight{1.25}\smash{\begin{tabular}[t]{l}0\end{tabular}}}}%
    \put(0.06123065,0.34306025){\makebox(0,0)[lt]{\lineheight{1.25}\smash{\begin{tabular}[t]{l}100\end{tabular}}}}%
    \put(0.03895546,0.118443){\rotatebox{90}{\makebox(0,0)[lt]{\lineheight{1.25}\smash{\begin{tabular}[t]{l}torque (in \si{Nm})\end{tabular}}}}}%
    \put(0.51698671,0.36860361){\makebox(0,0)[lt]{\lineheight{1.25}\smash{\begin{tabular}[t]{l}\textbf{knee}\end{tabular}}}}%
    \put(0,0){\includegraphics[width=\unitlength,page=3]{plot_InterSubject_torqueEstimation_knee_p6547_s60.pdf}}%
  \end{picture}%
\endgroup%

%% file: figures/plot_IntraSubject_torqueEstimation_knee_p6547_s60.pdf_tex
\begingroup%
  \makeatletter%
  \providecommand\color[2][]{%
    \errmessage{(Inkscape) Color is used for the text in Inkscape, but the package 'color.sty' is not loaded}%
    \renewcommand\color[2][]{}%
  }%
  \providecommand\transparent[1]{%
    \errmessage{(Inkscape) Transparency is used (non-zero) for the text in Inkscape, but the package 'transparent.sty' is not loaded}%
    \renewcommand\transparent[1]{}%
  }%
  \providecommand\rotatebox[2]{#2}%
  \newcommand*\fsize{\dimexpr\f@size pt\relax}%
  \newcommand*\lineheight[1]{\fontsize{\fsize}{#1\fsize}\selectfont}%
  \ifx\svgwidth\undefined%
    \setlength{\unitlength}{750bp}%
    \ifx\svgscale\undefined%
      \relax%
    \else%
      \setlength{\unitlength}{\unitlength * \real{\svgscale}}%
    \fi%
  \else%
    \setlength{\unitlength}{\svgwidth}%
  \fi%
  \global\let\svgwidth\undefined%
  \global\let\svgscale\undefined%
  \makeatother%
  \begin{picture}(1,0.4)%
    \lineheight{1}%
    \setlength\tabcolsep{0pt}%
    \put(0,0){\includegraphics[width=\unitlength,page=1]{plot_IntraSubject_torqueEstimation_knee_p6547_s60.pdf}}%
    \put(0.16167025,0.05409355){\makebox(0,0)[lt]{\lineheight{1.25}\smash{\begin{tabular}[t]{l}0\end{tabular}}}}%
    \put(0.34500965,0.05409355){\makebox(0,0)[lt]{\lineheight{1.25}\smash{\begin{tabular}[t]{l}4\end{tabular}}}}%
    \put(0.52834895,0.05409355){\makebox(0,0)[lt]{\lineheight{1.25}\smash{\begin{tabular}[t]{l}8\end{tabular}}}}%
    \put(0.70518835,0.05409355){\makebox(0,0)[lt]{\lineheight{1.25}\smash{\begin{tabular}[t]{l}12\end{tabular}}}}%
    \put(0.88852765,0.05409355){\makebox(0,0)[lt]{\lineheight{1.25}\smash{\begin{tabular}[t]{l}16\end{tabular}}}}%
    \put(0.48814268,0.02088707){\makebox(0,0)[lt]{\lineheight{1.25}\smash{\begin{tabular}[t]{l}time (in \si{\s})\end{tabular}}}}%
    \put(0,0){\includegraphics[width=\unitlength,page=2]{plot_IntraSubject_torqueEstimation_knee_p6547_s60.pdf}}%
    \put(0.04923065,0.13026025){\makebox(0,0)[lt]{\lineheight{1.25}\smash{\begin{tabular}[t]{l}-100\end{tabular}}}}%
    \put(0.09423065,0.23666025){\makebox(0,0)[lt]{\lineheight{1.25}\smash{\begin{tabular}[t]{l}0\end{tabular}}}}%
    \put(0.06123065,0.34306025){\makebox(0,0)[lt]{\lineheight{1.25}\smash{\begin{tabular}[t]{l}100\end{tabular}}}}%
    \put(0.03895546,0.118443){\rotatebox{90}{\makebox(0,0)[lt]{\lineheight{1.25}\smash{\begin{tabular}[t]{l}torque (in \si{Nm})\end{tabular}}}}}%
    \put(0.51698671,0.36860361){\makebox(0,0)[lt]{\lineheight{1.25}\smash{\begin{tabular}[t]{l}\textbf{knee}\end{tabular}}}}%
    \put(0,0){\includegraphics[width=\unitlength,page=3]{plot_IntraSubject_torqueEstimation_knee_p6547_s60.pdf}}%
  \end{picture}%
\endgroup%

%% file: sections/discussion.tex
\section{DISCUSSION}
\label{sec:discussion}

This work presents a \gls{FMG}-based approach to estimating torques in the knee and ankle joints. It uses joint angles and velocities together with muscle activity in a \gls{GPR}. The effectiveness of this FMG-based method is validated through a study with ten participants and compared to a baseline model using only joint angle and velocity, as well as a model based on \gls{EMG}.

The \gls{LOPO} validation showed that a model based on joint angle and velocity is more effective for generalization across multiple users, including novel ones.
This observation may be due to the comprehensive recording of the full range of motion in each experiment, resulting in joint angle and velocity data that are more consistent across participants.
In contrast, muscle activity is influenced by factors such as varying fitness levels and fatigue, resulting in fluctuations in amplitudes even after calibration.
The results suggest that the electrical activation of muscles tends to be more uniform across individuals, as shown by the similar \gls{RMSE} of the \gls{EMG} signals in the \gls{LOPO} validation.
In contrast, the \gls{RMSE} of the \gls{LOPO} validation suggests that the volume of muscle activity itself is less consistent across individuals.
These results may be attributed to the validation of isokinetic motion focused on the sagittal plane. 
Previous research has shown that the integration of muscle activity into joint torque estimation models is more advantageous for non-cyclic than for cyclic tasks~\cite{scherpereel_improving_2024}.
Therefore, future research will include a range of activities of daily living to confirm the findings across a spectrum of combined motions and to determine the comprehensibility of muscular signals during motions, including multidirectional and non-periodic motions. 

The integration of muscle activity into the joint torque estimation appears to improve generalization to novel task characteristics, such as changes in task velocity and variations in the torque amplitude.
This is reflected in the low \gls{RMSE} values observed in the \gls{LOVO} validation for several participants and is supported by the visualization of the corresponding torque trajectories of the five exemplary swings. 
In this context, the estimation based on \gls{FMG} demonstrates greater accuracy than the \gls{EMG}-based model, which is almost comparable to the baseline model.
Belyea~\etal~\cite{belyea_fmg_2019} reported similar findings in their study of human wrist motion utilizing \gls{SVR}.

The proposed \gls{GPR} model uses a fully data-driven approach with a difficult-to-interpret model.
Future research will aim to integrate additional prior knowledge into the \gls{GPR} model, drawing inspiration from biomechanical models or exploring other more interpretable model alternatives.
This has the potential to reduce the data requirements for model training and provide a clearer understanding of how each input signal influences the model output.
In addition, it may be possible to distinguish between a generalizable model based on joint angles and velocities and a customizable model based on muscle activity. 
This separation could allow the general model to be optimized for each individual during use.
In contrast, less comprehensive and more data-driven models such as \gls{ANN}~\cite{zhang_ankle_2021} and \gls{CNN}~\cite{schulte_multi-day_2022} have so far proven to be well suited for \gls{EMG}-based joint torque estimation in isokinetic and non-weight-bearing applications as well as during walking. 

In the future, when using an exoskeleton, it is expected that the forces used to move the exoskeleton may interfere with the measurement of muscle activity using \gls{FMG} due to potential disturbance forces caused by the interaction between the exoskeleton and the user. 
Even within this laboratory setup, it remains a challenge to eliminate the possibility that other parasitic forces - arising from co-contraction or external perturbations - may interfere with \gls{FMG} measurements. 
The decision to use the IsoMed device was aimed at mitigating the influence of as many parasitic forces as possible. 
In our previous work, we have shown how motion restrictions due to an ankle exoskeleton can affect \gls{FMG} sensor signals~\cite{Marquardt2024}. 
Further research will reveal the extent to which actuation can affect \gls{FMG}-based torque estimation and control.

The results emphasize that \gls{FMG} technology offers a promising alternative to \gls{EMG} technology for the development of a more personalized assistive exoskeleton, but further research on its performance and applicability in exoskeleton control is needed.

%% file: sections/conclusion.tex
\section{CONCLUSION}
\label{sec:conclusion}

Incorporating muscle activity signals is a promising approach for online personalization of exoskeleton control. In this work, we proposed an \gls{FMG}-based approach to estimate knee and ankle joint torques using joint angles, velocities, and muscle activity in a \gls{GPR} model. 
To validate the effectiveness of \gls{FMG}-based torque estimation, we conducted a user study with ten subjects performing isokinetic ankle and knee motion.
A comparative analysis was performed against a baseline model using only joint angle and velocity data, and a model augmented with muscle activity signals obtained from \gls{EMG} sensors.
The results reveal that incorporating \gls{FMG} data into the exoskeleton control improves the estimation of joint torque for the ankle and knee joints in the presence of unknown task characteristics, such as velocity changes, for each subject.
While the results may not show improved generalizability across different subjects, the findings underscore the need for further research into the potential of FMG as an alternative to EMG for developing control strategies for exoskeletons.